\documentclass[lettersize,journal]{IEEEtran}
\usepackage{multirow}
\usepackage{amsmath}
\usepackage{amsfonts}
\usepackage{amssymb}
\usepackage{array}
\usepackage[caption=false,font=normalsize,labelfont=sf,textfont=sf]{subfig}
\usepackage{textcomp}
\usepackage{stfloats}
\usepackage{url}
\usepackage{makecell}
\usepackage{verbatim}
\usepackage{graphicx}
\usepackage{color}
\usepackage{cite}
\usepackage{bbding}
\usepackage{amssymb}
\usepackage{algpseudocode}
\usepackage[utf8]{inputenc}
\usepackage{algorithmicx}
\usepackage{algpseudocode}
\usepackage{algorithm}
\usepackage{booktabs}
\DeclareUnicodeCharacter{FF1A}{\textasciitilde}
\usepackage{makecell}
\usepackage[colorlinks,
            linkcolor=red,
            anchorcolor=blue,
            citecolor=green 
            ]{hyperref}

\begin{document}

\title{Rethinking Generalizable Infrared Small Target Detection: A Real-scene Benchmark and Cross-view Representation Learning}

\author{Yahao Lu, Yuehui Li, Xingyuan Guo, Shuai Yuan, Yukai Shi, Liang Lin,~\IEEEmembership{Fellow,~IEEE,}
        % <-this % stops a space
\thanks{ 
%This work was supported in part by National Natural Science Foundation (62373112), in part by  Guangzhou Key R\&D Program (202206010104), in part by National Key R\&D Program of China (no.2021ZD0111600) and Natural Science Foundation of Guangdong Province (2021A1515011141).

Y. Lu, Y Li and Y. Shi are with School of Information Engineering, Guangdong University of Technology, Guangzhou, 510006, China (email: 2112303120@mail2.gdut.edu.cn; 3222000092@mail2.gdut.edu.cn; ykshi@gdut.edu.cn). 

X. Guo is with Southern Power Grid, Ltd., Guangzhou, 510000, China. (email: { gxypower@foxmail.com}).

Shuai Yuan is with the Xi'an Key Laboratory of Infrared Technology and System, School of Optoelectronic Engineering, Xidian University, Xi'an 710071, China. (email: yuansy@stu.xidian.edu.cn)

L. Lin is with School of Data and Computer Science, Sun Yat-sen University, Guangzhou, 510006, China (email: linliang@ieee.org).
}}


\maketitle

\begin{abstract}
Infrared small target detection (ISTD) is highly sensitive to sensor type, observation conditions, and the intrinsic properties of the target. These factors can introduce substantial variations in the distribution of acquired infrared image data, a phenomenon known as domain shift. Such distribution discrepancies significantly hinder the generalization capability of ISTD models across diverse scenarios. To tackle this challenge, this paper introduces an ISTD framework enhanced by domain adaptation. To alleviate distribution shift between datasets and achieve cross-sample alignment, we introduce Cross-view Channel Alignment (CCA). Additionally, we propose the Cross-view Top-K Fusion strategy, which integrates target information with diverse background features, enhancing the model’s ability to extract critical data characteristics. To further mitigate the impact of noise on ISTD, we develop a Noise-guided Representation learning strategy. This approach enables the model to learn more noise-resistant feature representations, to improve its generalization capability across diverse noisy domains. Finally, we develope a dedicated infrared small target dataset, RealScene-ISTD. Compared to state-of-the-art methods, our approach demonstrates superior performance in terms of detection probability ($P_d$), false alarm rate ($F_a$), and intersection over union ($IoU$). The code is available at: \url{https://github.com/luy0222/RealScene-ISTD}.
\end{abstract}

\begin{IEEEkeywords}
Infrared Small Target Detection, Channel Alignment, Cross-view Representation Learning, Top-K Fusion.
\end{IEEEkeywords}

\begin{figure}[ht]
    \centering
    \includegraphics[width=0.45\textwidth]{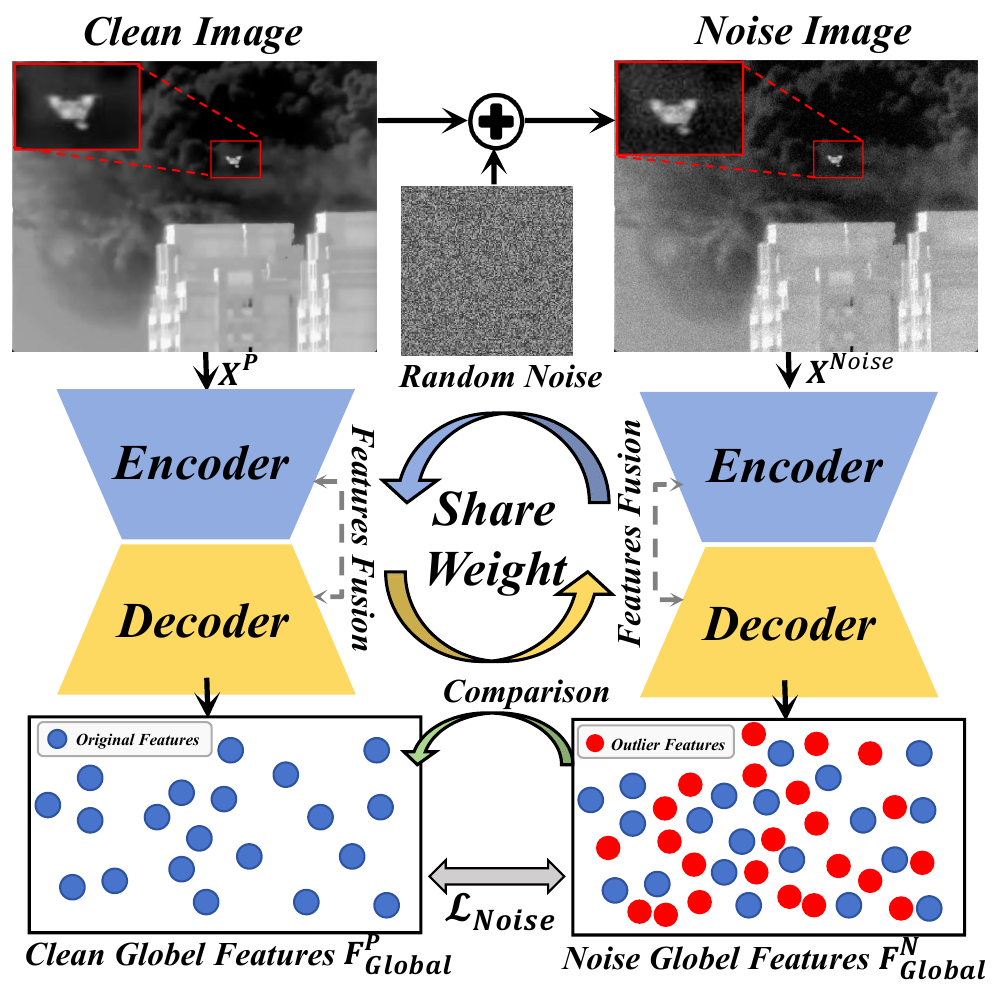}
    \caption[something short]{Our Noise-guided Representation  Learning process. Compared to traditional ISTD work, we generate outlier features (\textcolor{red}{red dots}) in a noise-guided manner. By minimizing the feature space distance between $F^P_{Global}$ and $F^N_{Global}$, we guide the model to learn more essential feature representations from limited data (\textcolor{blue}{blue dots}). }
    \label{fig1}
\end{figure}

\section{Introduction}
\IEEEPARstart{I}{STD} -technology is gradually expanding into domains such as security surveillance~\cite{security1,security2}, search and rescue~\cite{rescue} operations, autonomous driving~\cite{autonomous}, and environmental monitoring~\cite{monitoring1,monitoring2}, demonstrating significant application value. For the effective detection of small targets in infrared images, researchers have proposed various methods. These methods can be divided into two categories: traditional methods~\cite{MITHF,Mclntosh,WSLCM,TLLCM,NRAM} and deep learning methods~\cite{deep2,deep1,deep3}. Early ISTD methods primarily relied on traditional techniques such as image filtering~\cite{filtering1,filtering2,filtering3} and human visual system (HVS)~\cite{HVS1,HVS2} mechanisms. These methods often rely on intricate handcrafted feature designs, which limit their adaptability to the complex and dynamically evolving real-world scenarios.

With the significant advancements in hardware computing power, deep learning-based ISTD methods have garnered increasing attention and popularity. To more effectively capture the fine contours of targets and mitigate performance degradation caused by small target sizes, researchers continue to explore increasingly sophisticated model architectures. Dai et al. proposed a local contrast network (ALCNet)~\cite{ALC}  based on an attention mechanism. This network enhances small target detection by effectively merging local contrast with low-level features, leading to improved detection rates. To further enhance feature representation, Li et al. proposed the Dense Nested Attention Network (DNA-Net)~\cite{DNA}, which achieves progressive enhancement of targets by iteratively fusing feature maps of different scales. With the breakthrough progress of Transformer models in the field of computer vision and the rapid improvement in hardware computing power, data-driven feature extraction methods have gained increasing attention due to their strong representation learning capabilities. Yuan et al. integrated the Transformer~\cite{vaswani2017attention} architecture with a spatial-channel cross-attention mechanism into the classical U-Net~\cite{unet} network, establishing image-scale information associations through semantic interactions at multiple levels, significantly boosting model performance and reaching the current optimal level. However, existing research primarily focuses on enhancing the detection performance, while overlooking the shift of real-world data distribution:
\begin{itemize}
    
    \item[$\bullet$] \textbf{Out-of-distribution (o.o.d.) challenges}: In real-world applications, training and testing image data often exhibit distribution shift, due to the variations in equipment, temperature, humidity, lighting, and the collection scene.
    
    \item[$\bullet$] \textbf{Algorithm robustness}: The ISTD process is highly sensitive to variations in sensors, lighting conditions, and target diversity, which can lead to a decline in model accuracy and make target detection more challenging.
\end{itemize}

\begin{center}
    \begin{table*}[!t]
    \caption{Comparison of State-of-the-Art ISTD Methods in Multi-Dataset Scenarios. Red highlights indicate the performance drop in multi-dataset scenarios compared to single-dataset scenarios.}
    \label{table1}
    \setlength{\tabcolsep}{5pt}
    \small
    \centering
    \resizebox{2\columnwidth}{!}{
% \usepackage{multirow}

  \begin{tabular}{|c|ccc|ccc|cccccc|}
    \hline
    \multirow{3}{*}{Method} & \multicolumn{3}{c|}{Training Set:  IRSTD-1K}            & \multicolumn{3}{c|}{Training Set:  RealScene-ISTD}                & \multicolumn{6}{c|}{Training Set: NUAA-SIRST, IRSTD-1K, RealScene-ISTD}                                                                                                                                                                                                                                                                                                                                                                                                  \\  \cline{2-13} 
                       & \multicolumn{3}{c|}{Test on IRSTD-1K}                           & \multicolumn{3}{c|}{Test on RealScene-ISTD}                      & \multicolumn{3}{c}{Test on IRSTD-1K}                                                                                                                                                                                                             & \multicolumn{3}{c|}{Test on RealScene-ISTD}                                                                                                                                                                                 \\ \cline{2-13} 
                       & \multicolumn{1}{c}{$IoU\uparrow$}  & \multicolumn{1}{c }{$P_d\uparrow$}   & $F_a\downarrow$   & \multicolumn{1}{c}{$IoU\uparrow$}  & \multicolumn{1}{c }{$P_d\uparrow$}   & $F_a\downarrow$    & \multicolumn{1}{c}{$IoU\uparrow$}                                                      & \multicolumn{1}{c}{$P_d\uparrow$}                                                     & \multicolumn{1}{c}{$F_a\downarrow$}                                                          & \multicolumn{1}{c}{$IoU\uparrow$}                                                     & \multicolumn{1}{c}{$P_d\uparrow$}                                                     & $F_a\downarrow$                                                          \\ \hline \hline   
    ACM~\cite{ACM}                & \multicolumn{1}{c }{59.41} & \multicolumn{1}{c }{93.96} & 84.36 & \multicolumn{1}{c }{66.26} & \multicolumn{1}{c }{91.54} & 190.59 & \multicolumn{1}{c }{\begin{tabular}[c]{@{}c@{}}57.68\\ \color{red}{ ($\downarrow$1.73\%)}\end{tabular}} & \multicolumn{1}{c }{\begin{tabular}[c]{@{}c@{}}92.23\\ \color{red}{($\downarrow$1.73\%)}\end{tabular}} & \multicolumn{1}{c }{\begin{tabular}[c]{@{}c@{}}115.28\\ \color{red}{($\downarrow$30.92\%)}\end{tabular}}    & \multicolumn{1}{c }{\begin{tabular}[c]{@{}c@{}}64.25\\ \color{red}{($\downarrow$2.01\%)}\end{tabular}}  & \multicolumn{1}{c }{91.97}                                                   & \begin{tabular}[c]{@{}c@{}}250.79\\\color{red}{ ($\downarrow$60.11\%)}\end{tabular}    \\ \hline
    DNA-Net~\cite{DNA}                & \multicolumn{1}{c }{65.03} & \multicolumn{1}{c }{94.30} & 18.83 & \multicolumn{1}{c }{75.02} & \multicolumn{1}{c }{91.75} & 81.75  & \multicolumn{1}{c }{\begin{tabular}[c]{@{}c@{}}64.14\\\color{red}{ ($\downarrow$0.89\%)}\end{tabular}}   & \multicolumn{1}{c }{95.64}                                                   & \multicolumn{1}{c }{\begin{tabular}[c]{@{}c@{}}40.90\\ \color{red}{($\downarrow$22.07\%)}\end{tabular}}     & \multicolumn{1}{c }{\begin{tabular}[c]{@{}c@{}}73.79\\ \color{red}{($\downarrow$1.23\%)}\end{tabular}}  & \multicolumn{1}{c }{\begin{tabular}[c]{@{}c@{}}91.33\\ \color{red}{($\downarrow$0.42\%)}\end{tabular}} & \begin{tabular}[c]{@{}c@{}}60.06\\ \color{red}{($\downarrow$21.69\%)}\end{tabular}     \\ \hline
    RDIAN~\cite{RDIAN}                & \multicolumn{1}{c }{63.95} & \multicolumn{1}{c }{94.30} & 47.07 & \multicolumn{1}{c }{63.37} & \multicolumn{1}{c }{87.74} & 171.45 & \multicolumn{1}{c }{\begin{tabular}[c]{@{}c@{}}60.12\\ \color{red}{ ($\downarrow$3.83\%)}\end{tabular}}   & \multicolumn{1}{c }{\begin{tabular}[c]{@{}c@{}}93.96\\ \color{red}{($\downarrow$0.34\%)}\end{tabular}} & \multicolumn{1}{c }{\begin{tabular}[c]{@{}c@{}}98.88\\ \color{red}{ ($\downarrow$51.81\%)}\end{tabular}}     & \multicolumn{1}{c }{\begin{tabular}[c]{@{}c@{}}59.46\\ \color{red}{($\downarrow$3.91\%)}\end{tabular}}  & \multicolumn{1}{c }{90.70}                                                   & \begin{tabular}[c]{@{}c@{}}357.92\\ \color{red}{($\downarrow$186.47\%)}\end{tabular}   \\  \hline
    UIU-Net~\cite{UIU}               & \multicolumn{1}{c }{63.25} & \multicolumn{1}{c }{95.30} & 61.94 & \multicolumn{1}{c }{70.03} & \multicolumn{1}{c }{89.01} & 102.95 & \multicolumn{1}{c }{\begin{tabular}[c]{@{}c@{}}39.10\\ \color{red}{ ($\downarrow$24.15\%)}\end{tabular}}  & \multicolumn{1}{c }{\begin{tabular}[c]{@{}c@{}}92.62\\ \color{red}{($\downarrow$2.68\%)}\end{tabular}} & \multicolumn{1}{c}{\begin{tabular}[c]{@{}c@{}}7008.86\\ \color{red}{ ($\downarrow$6946.92\%)}\end{tabular}} & \multicolumn{1}{c }{\begin{tabular}[c]{@{}c@{}}31.75\\ \color{red}{($\downarrow$38.28\%)}\end{tabular}} & \multicolumn{1}{c }{\begin{tabular}[c]{@{}c@{}}89.43\\ \color{red}{($\downarrow$0.42\%)} \end{tabular}} & \begin{tabular}[c]{@{}c@{}}2928.64\\ \color{red}{($\downarrow$2825.69\%)}\end{tabular} \\ \hline
    SCTransNet~\cite{sctransnet}             & \multicolumn{1}{c }{64.11} & \multicolumn{1}{c }{94.97} & 16.68 & \multicolumn{1}{c }{76.56} & \multicolumn{1}{c }{93.66} & 47.19  & \multicolumn{1}{c }{66.93}                                                     & \multicolumn{1}{c }{\begin{tabular}[c]{@{}c@{}}93.96\\ \color{red}{($\downarrow$1.01\%)}\end{tabular}}                                                 & \multicolumn{1}{c }{15.45}                                                        & \multicolumn{1}{c }{77.00}                                                    & \begin{tabular}[c]{@{}c@{}}93.23\\ \color{red}{($\downarrow$0.43\%)}\end{tabular}                 & \begin{tabular}[c]{@{}c@{}}60.51\\ \color{red}{($\downarrow$13.32\%)} \end{tabular}     \\ \hline
    \end{tabular}
    }
    \label{tab1}
    \end{table*}
\end{center}

The limited generalization capability of the model severely restricts its applicability in complex and dynamic real-world scenarios.  Recently, researchers have explored diverse solutions, including sky-labeled~\cite{sky} data, Mosaic~\cite{mosaic} data augmentation, and energy-guided single-point prompting~\cite{Beyondfulllabels}, to improve model robustness against domain shifts. These methods primarily focus on structural features of targets under different backgrounds, such as shape, pose. However, they failed to take into account the effects arising from variations in data domain features. As shown in Tab.~\ref{table1}, when directly using diverse sensor-collected data, the diversity of dataset distribution reduces the model generalization ability. Moreover, relying solely on a single dataset for feature extraction can cause the model to fall into a local optimum, limiting its expressive power. Therefore, effectively addressing deep feature distribution discrepancies and data distribution mismatches is crucial for improving the generalization performance of ISTD models.

To alleviate domain discrepancies in ISTD models, we propose the Cross-view Dataset Alignment strategy, which enhances the domain adaptation capability of detection and improves model generalization across diverse and complex environments. Additionally, to uncover richer information within the dataset, we introduce a Top-K Fusion strategy. This approach utilizes SSIM similarity to match target objects with varied background images, generating training samples with richer variations. To further mitigate image domain shifts induced by infrared thermal noise, as illustrated in Fig.~\ref{fig1}, we present a Noise-guided Representation  Learning strategy. By using noise perturbations into input images during training, we simulate thermal radiation variations across different sensors, enabling the model to learn noise-resistant feature representations and better adapt to domain shifts in the ISTD process.
Additionally, we construct a cross-domain infrared small target dataset, RealScene-ISTD, designed to facilitate research on model generalization under cross-domain data conditions and to promote performance optimization of existing ISTD algorithms in real-world environments. As shown in Tab.~\ref{table2}, our method significantly enhances ISTD detection performance across various scenarios and extreme challenges.

In summary, the key contributions of this paper are as follows:
\begin{itemize}

    \item[$\bullet$] We introduce the Cross-view Dataset Alignment and Fusion strategy, which applies cross-sensor transformation and augmentation to existing datasets in a given target domain, effectively mitigating domain discrepancies.
    
    \item[$\bullet$] We propose the Noise-guided Representation  Learning strategy. This method can guide the model to learn noise-resistant features, enhancing the model’s generalization ability in noisy domains across sensors.
    
    \item[$\bullet$] We present RealScene-ISTD, a novel cross-domain infrared small target dataset featuring varied shapes, categories, poses, and sensor modalities from multiple sources. It serves as a benchmark for evaluating model generalization across domains and advancing the real-world performance of ISTD algorithms.
    
\end{itemize}

\begin{figure*}[!t]
\centering
\includegraphics[width=0.95\linewidth]{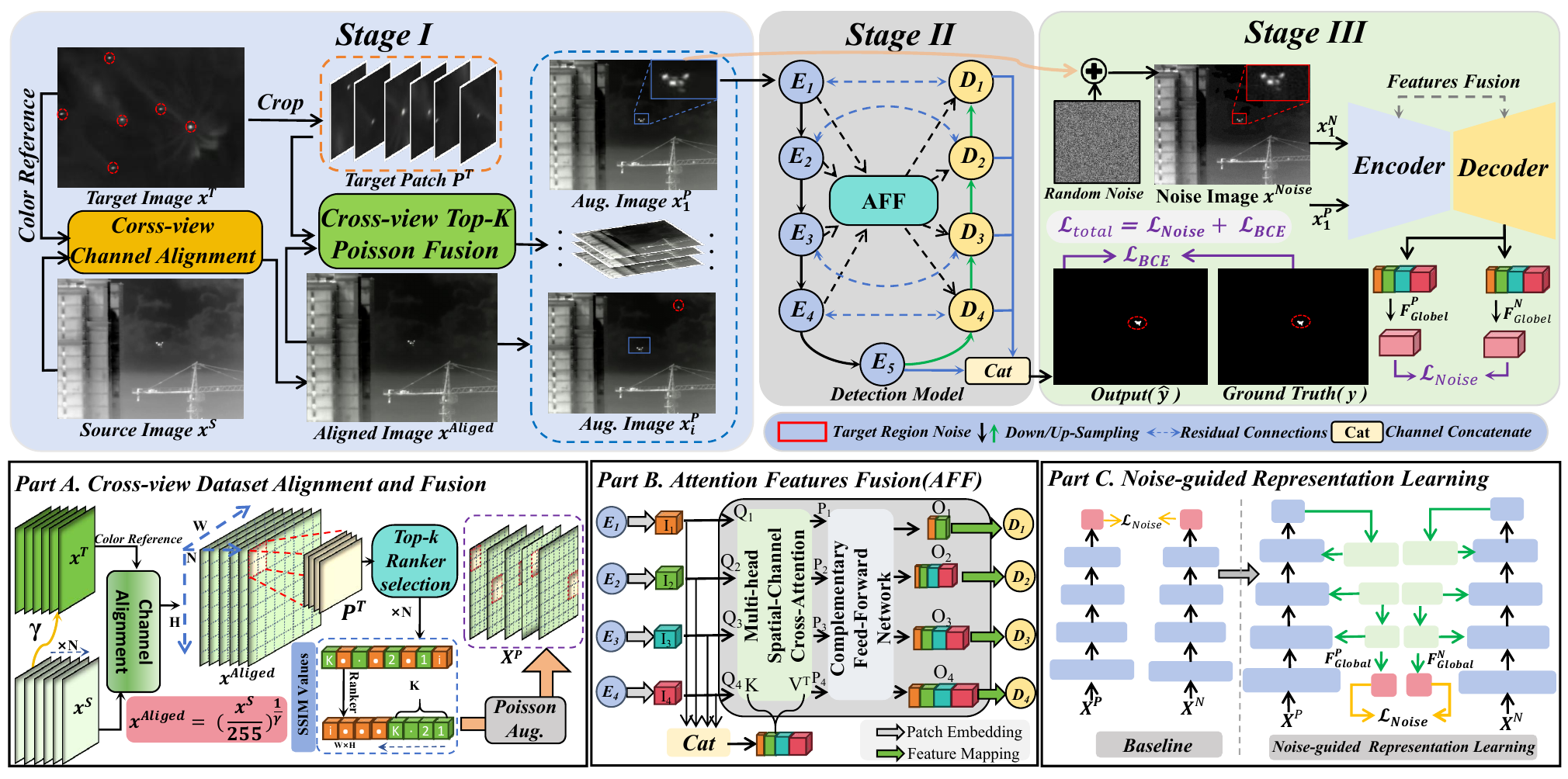}
\caption{Illustration of the proposed generalizable infrared small target detection framework. (a) The Cross-view Channel Alignment and Top-K Poisson Fusion method aims to effectively expand external data and deeply explore internal latent information to address domain shift issues; (b) The feature fusion module establishes semantic connections at different levels through multi-scale feature interactions; (c) The Noise-guided Representation  Learning strategy mitigates domain shift caused by noise variations in different infrared images, achieving higher generalization capability.}
\label{fig2}
\end{figure*}

\section{RELATED WORK}
\subsection{Single-frame Infrared Small Target Detection}
Infrared imaging is highly susceptible to variations in sensor types, observation conditions, and target characteristics, which limits generalization across diverse real-world scenarios. Consequently, accurately segmenting infrared small targets within  complex backgrounds remains a formidable challenge.

Traditional approaches, such as Tophat~\cite{Tophat}, LCM~\cite{LCM}, and IPI~\cite{IPI}, struggle to achieve robust generalization in dynamic and complex environments due to their lack of deep semantic understanding. In contrast, CNN-based models~\cite{cnn} exhibit stronger generalization capabilities by extracting high-level semantic features. For instance, MDvsFA-GAN~\cite{MDvsFA-GAN} employs generative adversarial networks for multimodal data transformation, UIU-Net~\cite{UIU} integrates global contextual information to enhance detection performance, and SCTransNet~\cite{sctransnet} leverages spatial and channel attention mechanisms to achieve state-of-the-art results.

Despite the significant advancements made by CNN-based methods, the substantial distribution discrepancies in infrared images and the limited availability of training data remain critical bottlenecks that hinder further performance improvements.

\subsection{Domain Generalization}
The effectiveness of deep learning models largely depends on the consistency of data distributions during training. However, in practical applications, discrepancies between different infrared small target datasets are common. These discrepancies can stem from various factors, including imbalanced data collection, labeling errors, and insufficient sample representativeness.

Domain adaptation\cite{DomainAdaptation1,DomainAdaptation2,DomainAdaptation3} mitigates dataset bias by mapping both source and target domain data into a shared feature space, enabling the model to better adapt to the target distribution. Domain generalization\cite{DG1,DG2,DG3} enhances model robustness by learning more generalized feature representations from multiple source domains, allowing for effective inference on unseen data. Test-time adaptation (TTA)~\cite{TTA1,TTA2,TTA3,TTA4} dynamically adjusts the model during inference, optimizing it in real time based on the characteristics of each sample. This improves the model’s adaptability to varying data distributions, enhancing both stability and generalization.

In real-world applications, ISTD models are particularly susceptible to domain shifts caused by differences in sensor types, observation environments, and target characteristics. Although current domain shift methods have shown some effectiveness in semantic segmentation\cite{ds1,ds2} and image fusion\cite{fusion}, the inherent challenges of the ISTD task, such as the diverse shapes and poses of small targets and the ubiquitous clutter interference in the background, make it difficult for these existing methods to be effectively applied to the ISTD domain. To address this, we propose the Cross-view Dataset Alignment strategy and the Noise-guided Representation  Learning strategy, designed to improve model robustness and generalization across diverse and complex environments.

\section{Methodology}
In this section, we elaborate on the implementation of Cross-view Channel Alignment, the Fusion-based Data Augmentation technique, and the Noise-guided Representation  Learning strategy. The overall framework of our proposed method is illustrated in Fig.~\ref{fig2}.

\subsection{Problem Definition} 
The inconsistency in training data distribution is a key factor limiting the model’s generalization capability. To illustrate the challenges caused  by distribution shift in ISTD, we train several representative ISTD methods using datasets with different distribution shifts and evaluate them on the same dataset. As shown in Tab.~\ref{table1}, the red arrows indicate the performance degradation of the model in multi-dataset scenarios compared to single-dataset ones. \emph{Given cross-domain training data, the state-of-the-art models meet a performance degradation, which restricts the generalization of ISTD models in a wide range of application scenes.}

\subsection{Cross-view Alignment and Fusion} \label{CCA}
To mitigate the domain discrepancies, we propose Cross-view Channel Alignment and Top-K Fusion. %This method enables effective domain adaptation without requiring additional training stages. 

\begin{figure}[t]
\centering
\includegraphics[width=0.95\columnwidth]{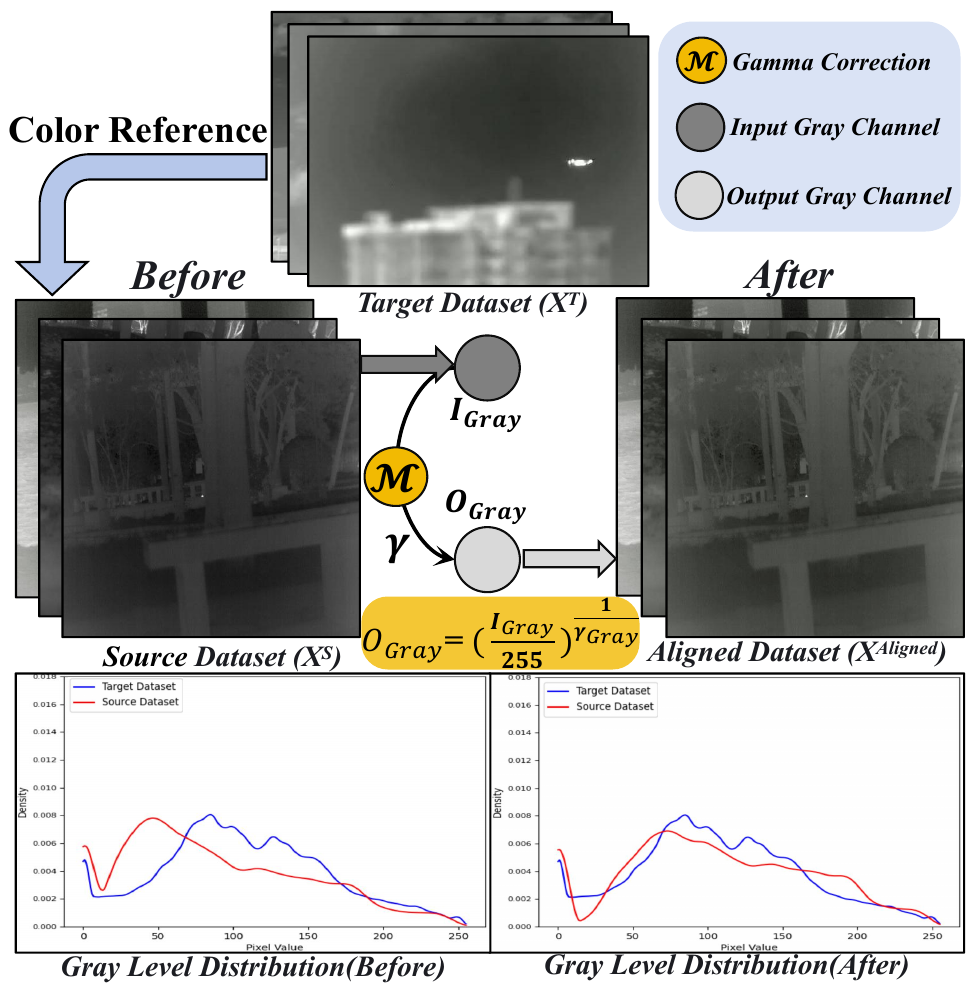}
\caption{Cross-view Channel Alignment effectively improves the alignment of average pixel distributions between source and target domain images, thereby significantly enhancing their visual similarity.} 
\label{fig3}
\end{figure}

\textbf{Cross-view Channel Alignment.} We define the source domain data as $X^S = \left \{ x^S_{u} \right \}_{u=1}^{N_S}$ and the target domain data as $X^T = \left \{ x^T_v \right \}_{v=1}^{N_T}$. In this context, $N_S$ and $N_T$ refer to the number of images in the source and target domains, respectively. We first calculate the channel gamma value $\gamma_{Gray}$ from the source domain to the target domain, and then perform gamma correction on the source domain image $X^{S}$. The formula is as follows:
\begin{equation} \label{eq1}
   O_{Gray}  = (\frac{I_{Gray}}{255} )^{\frac{1}{\gamma _{Gray} } },
\end{equation}
$O$ and $I$ represent the output and input domain pixel intensities, respectively, where $I\in[0,255]$. $\gamma$ is the gamma factor. The subscript ``Gray'' represents the grayscale channel, and the value of the channel is unique. The image data after gamma correction is represented as $X^{Aliged} = \left \{ x^{aliged}_u \right \}_{u=1}^{N_A}$, $N_A$ refer to the number of images after gamma correction. Specifically, Cross-view Channel Alignment effectively adapts to variations in various distribution shifts between the source and target domains, ensuring consistency in cross-dataset characteristics. As shown in Fig.~\ref{fig3}, after Cross-view Channel Alignment processing, the source domain dataset and the target domain dataset achieve better alignment in visual appearance. To intuitively illustrate our method, we present the algorithm flow of Cross-view Channel Alignment in Algorithm~\ref{alg:Channel Alignment}. %This processing effectively alleviates data shift between different datasets, thereby expanding the dataset size, enriching its diversity, and providing more sufficient data support for subsequent model training. 

\begin{figure}[t]
\centering
\includegraphics[width=0.99\linewidth]{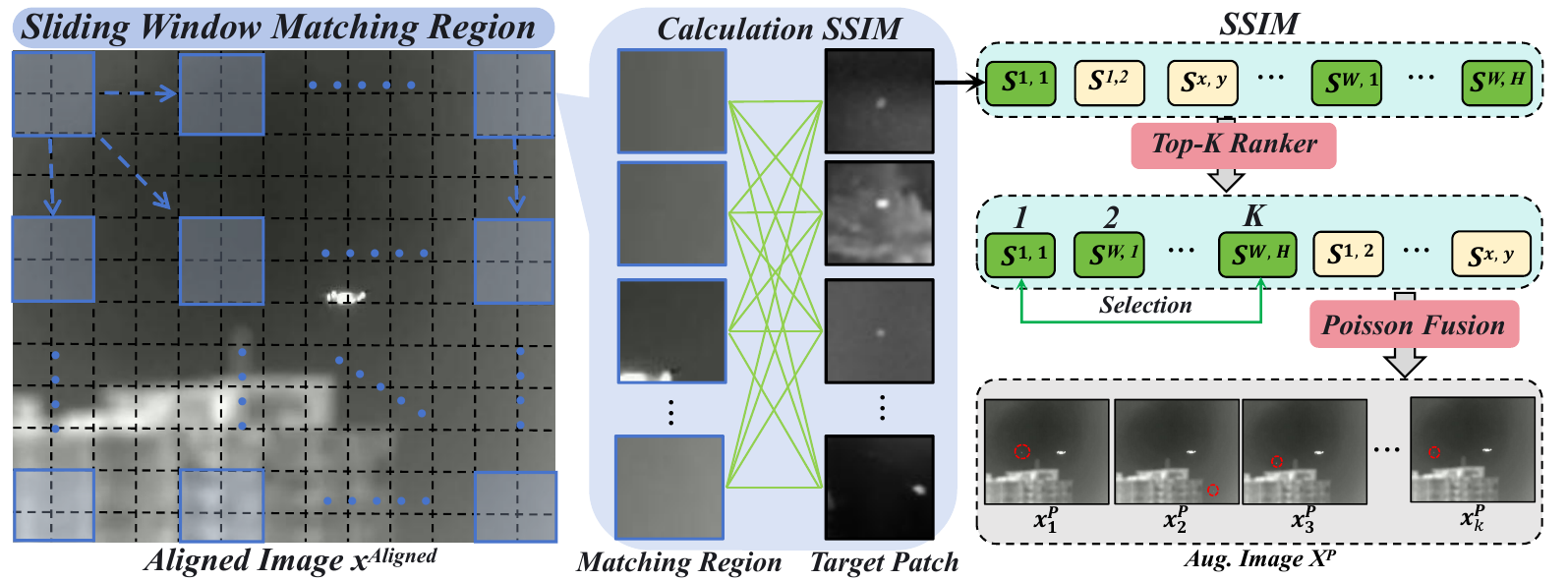}
\caption{Cross-view Top-K Poisson Fusion operates by scanning the image with a sliding window to identify local regions that closely match the target patch.. The top-matching regions are then selected and seamlessly blended using Poisson fusion. }
\label{fig4}
\end{figure}

 \textbf{Cross-view Top-K Fusion.} This process begins by extracting target patches from the image data $X^{Aliged}$, which has undergone Cross-view Channel Alignment, resulting in a large collection of target patches, denoted as $P^{T} = \left \{ p^{T}_n \right \}_{n=1}^{N}$. Next, fine-grained local contrast analysis is employed to precisely match highly similar regions between $X^{Aliged}$ and $P^{T}$. As illustrated in Fig.~\ref{fig4}, a sliding window approach is utilized to select matching regions across a complete image, and the Structural Similarity Index Measure (SSIM) between the selected region $Region^{x,y}$ and the target patch $P^{T}$ is computed.
\begin{equation}
    S^{x,y} = SSIM(Region^{x,y} , P^{T}),
\end{equation}
where $SSIM(\cdot)$ represents the calculation of the optimal SSIM index between the selected region $Region^{x,y}$ and the collection of target patches $P^{T}$. $S^{x,y}$ represents the optimal SSIM values computed over matching regions at different spatial locations for the target patch $ P^{T}$, where the superscript $x$,$y$ denotes the coordinates of the top-left corner of the matching region. We achieve a more natural blending effect and effectively reduce data redundancy by performing Poisson blending on the Top-K image patches with the smallest differences.

\begin{algorithm}[t!]
\caption{Cross-view Channel Alignment}
\label{alg:Channel Alignment}
\begin{algorithmic}

\State \textbf{Input:} Source images $X^{{S}}$, Target images $X^{{T}}$ 
\State \textbf{Intermediate parameter:} Gamma transform values  $\gamma_{Gray}$
 \State {\bf Output:} Aligned image $X^{{Aliged}}$
\\
 $\triangleright$ Compute channel average pixel intensity
\For{$u\gets1$ to $N_S$}
    \State $Mean(X^S)$ = $Mean(x^{s}_{u} )$
\EndFor
\For{$v\gets1$ to $N_T$}
    \State$Mean(X^T)$ =  $Mean(x^{t}_{v} )$
\EndFor

\State $\triangleright$ Get $\gamma _{Gary}$ and Gamma channel migration
\State $Mean(X^T)$ =$\left (\frac{Mean({X^S})}{255} \right )^{\frac{1}{\gamma _{Gary} } } $
\For{$u\gets1$ to $N_S$}
     \State $x^{aliged}_u$=$\left (\frac{x^{S}_u}{255} \right )^{\frac{1}{\gamma _{Gray} } }$
\EndFor

\end{algorithmic}
\end{algorithm}

\begin{algorithm}[t!]
\caption{Cross-view Top-K Poisson Fusion}
\label{alg:Top-K}
\begin{algorithmic}
\State \textbf{Input:} Aligned source images $X^{Aliged}$ 
\State \textbf{Intermediate parameter:} Sliding Window Matching $Region^{x,y} \in x^{Aliged} $, Cropping the target region $P^{T} \in X^{Aliged}$
 \State {\bf Output:} Cross-view Top-K Fusion $X^{{P}} $
\\ 
 $\triangleright$  Sliding Window Matching Region Selection
\For{$x\gets1$ to $W$}
    \For{$y\gets1$ to $H$}
        \State $S^{x,y}$ = SSIM($Region^{x,y}$ ,$P^{T}$)
    \EndFor
\EndFor
\State $Region_{Top-k}^{x,y}$ = Mapping($Top_k$($[S^{1,1},S^{1,2},\dots,S^{x,y}]$))
\\

 $\triangleright$  Top-K  Poisson Fusion
\For {$k\gets1$ to $K$}
    \State  $X^P$= Poisson($X^{Aliged}$, $P^T$, $Region_{Top-k}^{x,y}$)
\EndFor
\end{algorithmic}
\end{algorithm}

\begin{figure}[t]
\centering
\includegraphics[width=0.95\columnwidth]{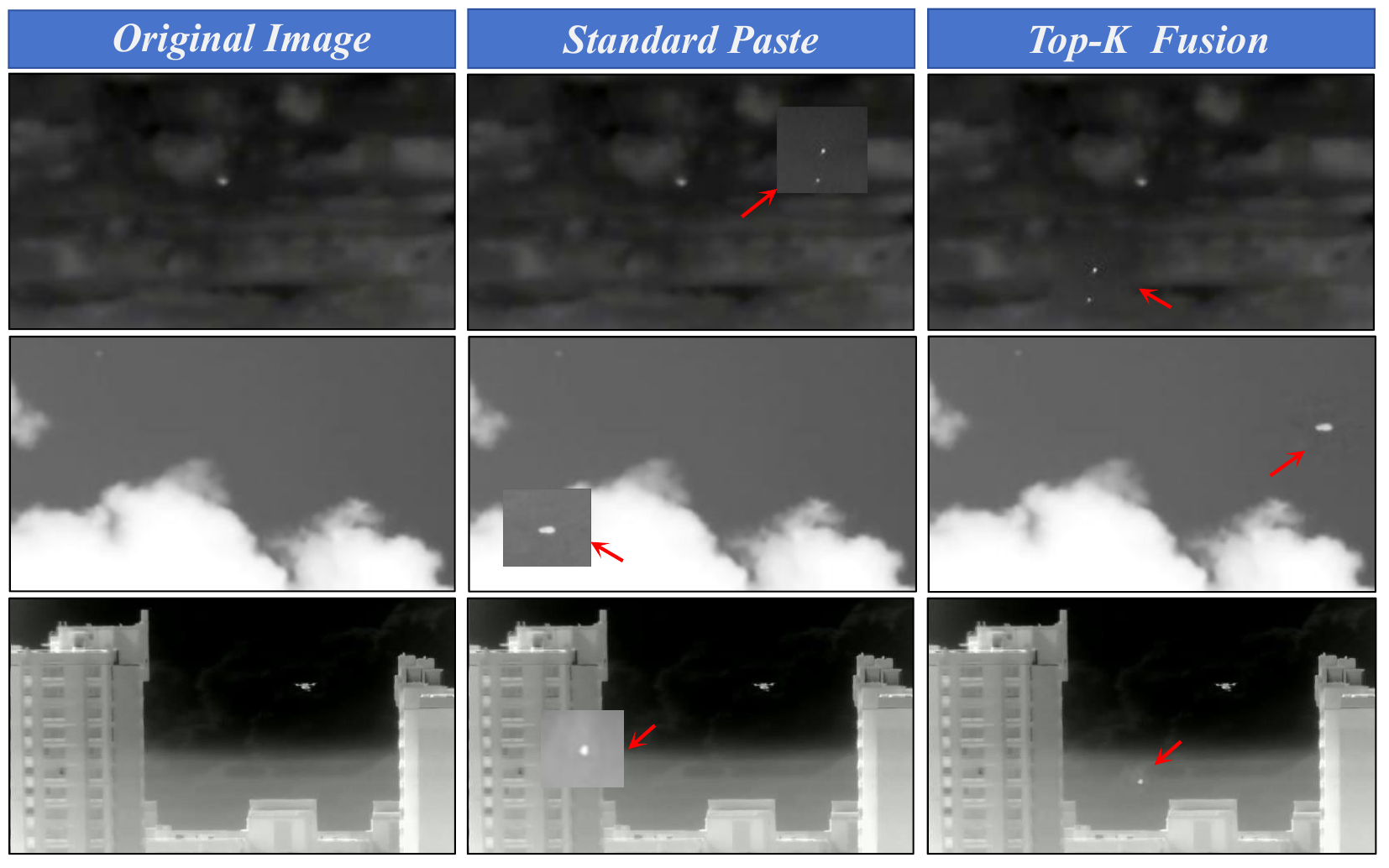}
\caption{Effectiveness of Cross-view Top-K Poisson Fusion. Unlike conventional image stitching techniques, Poisson fusion generates synthetic samples with greater visual authenticity.}
\label{fig5}
\end{figure}

\begin{equation}
    Region_{Top-k}^{x,y} = Mapping(Top_{k}([S^{1,1},S^{1,2},\dots,S^{x,y}])),
\end{equation}
where $Mapping(\cdot)$ denotes a mapping function that affixes $S^{x,y}$ to its corresponding positional information. $Region_{Top-k}^{x,y}$ represents the matching regions associated with the selected SSIM values, where $Top_k(\cdot)$ denotes the Top-K selection process. we perform Poisson fusion on the Top-K most similar regions, generating high-quality training samples $X^{P} = \left \{ x^{p}_{u} , y^{p}_{u}\right \}_{u=1}^{N_p}$ with diversified background-object compositions. The formulation is as follows:
\begin{equation}
    min\int_{(x,y)}\left \|  \bigtriangledown P^{T}  - \bigtriangledown {Region_{Top-k}^{x,y}}   \right \|  ^{2}  dxdy,
\end{equation}
In this manner, a large number of samples containing target patches $P^{T}$ embedded in diverse backgrounds are generated, enabling the model to better learn the intrinsic features of the images. Where $min\int_{(x,y)}\left\|\cdot\right \|^{2}dxdy$ represents the minimization integral, and  $\bigtriangledown$ denotes the gradient vector. The above formula, used as a constraint, effectively ensures the smoothness and realisticness. By applying Poisson fusion to the Top-K image patches with the smallest differences, we achieve a more seamless and natural fusion effect while effectively reducing data redundancy. The Top-K selection strategy optimizes computational efficiency and enhances training performance without significantly compromising model accuracy. As shown in Fig.~\ref{fig5}, the synthetic samples $X^{P} = \left \{ x^{p}_{u} , y^{p}_{u}\right \}_{u=1}^{N_p}$ generated by the Cross-view Top-K Fusion technique exhibit high visual realism, making it hard to distinguish them from real images. In this manner, a large number of samples containing target patches $P^{T}$ embedded in diverse backgrounds are generated, enabling the model to better learn the intrinsic features of the images. The implementation details are depicted in Algorithm~\ref{alg:Top-K}.

\subsection{Noise-guided Representation Learning} 
To further enhance the generalization performance of the ISTD model in cross-domain environments, we propose a Noise-guided Representation learning method.

\textbf{Feature Extraction. }  As shown in Fig.~\ref{fig2}, we adopt a network model $f(\cdot)$ based on the U-Net framework with Attention Features Fusion (AFF), which is used to extract infrared small target image features. $f(\cdot)$ consists of five layers of encoders and decoders. The encoder captures high-level features $F_i$ through a downsampling module with residual connections, and then the patches of $F_i$ are  ed to obtain the embedding layer $I_i$.  $i$ represents the layer number of the $f(\cdot)$ network. Then, $I_{i}$ undergoes full-level semantic feature fusion to obtain the output multi-scale feature $O_{i}$. We train the model $f(\cdot)$ to obtain the parameters $\theta$:
\begin{equation}
    \theta= argmin_{\theta}\frac{1}{{N_p} }\sum_{u=1}^{N_p}\mathcal{L}_{BCE}(f(x^{p}_{u};\theta),y^{p}_{u}),
\end{equation}
where $\mathcal{L}_{BCE}(\cdot)$ represents the Binary Cross-Entropy loss function computation. $x^{p}_{u}$ and $y^{p}_{u}$ represent the labeled Top-k Poisson Fusion data generated in Section~\ref{CCA}. $N_p$ indicates the number of samples in $X^P$.

\begin{figure}[t]
\centering
\includegraphics[width=0.90\columnwidth]{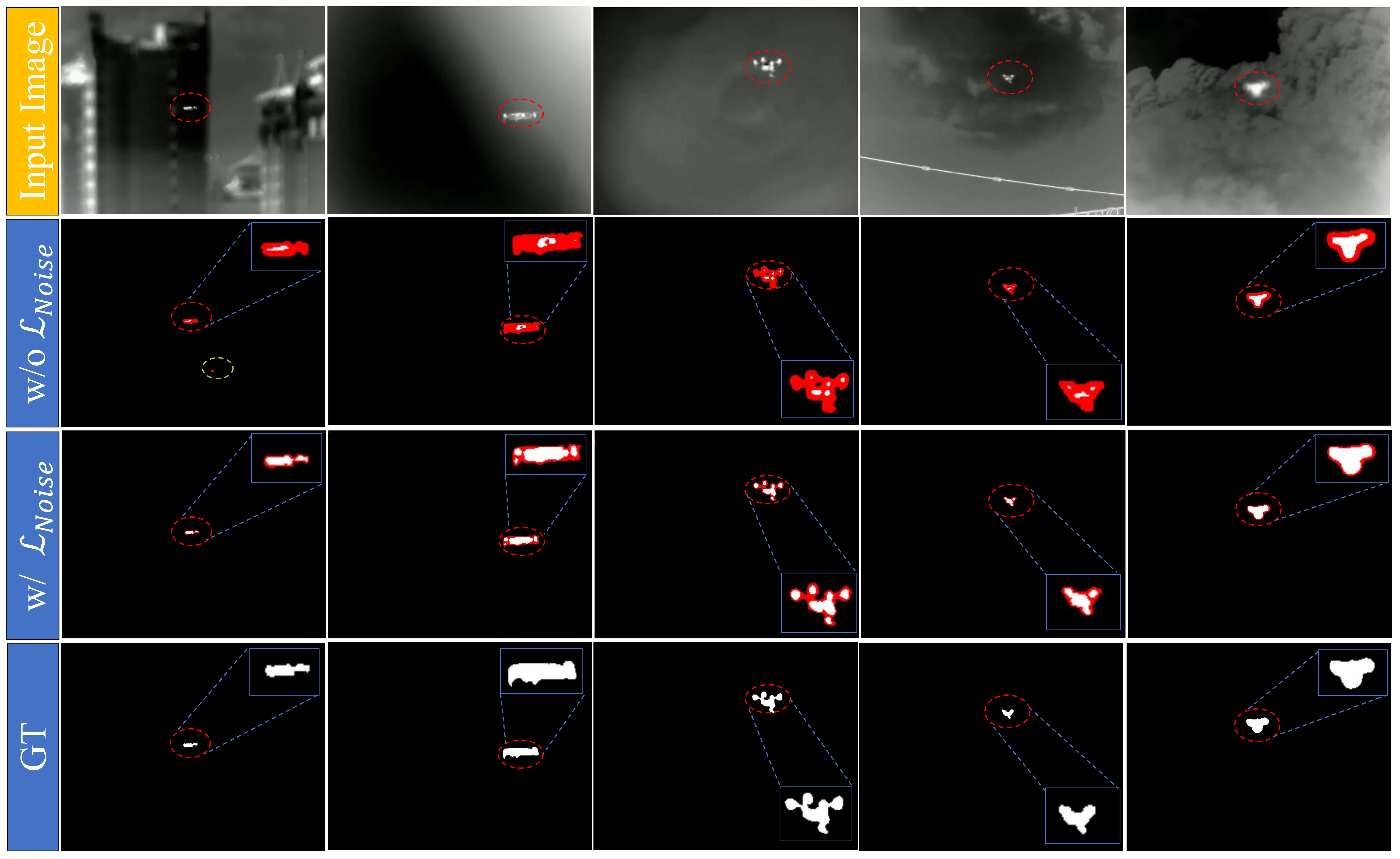}
\caption{Illustrate the impact of Noise-guided Representation  Learning $(\mathcal{L}_{Noise})$  on detection performance. In the visualization, \textcolor{red}{red pixels} highlight discrepancies between predictions and GT.}
\label{fig6}
\end{figure}

\textbf{Noise-guided Representation Learning.} As shown in Fig.~\ref{fig1}, we introduce random noise to implement a representation learning: 
\begin{equation}
    X^{Noise} = X^{P} + N(0,\alpha^2),
\end{equation}
where $N(0,\alpha^2)$ represents hermal noise that follows a distribution with a mean of $0$ and a variance of $\alpha^2$. $\alpha$ is a hyperparameter. $X^P$ represents generalized samples generated through Cross-view Top-K Fusion.  $X^{Noise}$ denotes the noise-added sample. We feed this into the shared weight model $f'(\cdot)$. By applying a feature consistency constraint, we guide the model to learn a unified representation of high-level features, enabling it to acquire general features that are robust to noise. To effectively extract and integrate features at various scales, we use a progressive downsampling strategy.  This strategy performs multi-scale fusion of the low-level and high-level features from both clean and noisy images, yielding the global features $F_{Global}^{P}$ and $F_{Global}^{N}$.
\begin{equation}
    F_{Global}^{P,N} = \left \{F_1^{P,N},  F_2^{P,N}, F_3^{P,N}, F_4^{P,N} \right \},
\end{equation}
where $F_i^{P}$ and $F_i^{N}$ refer to the i-th layer feature maps of the clean and noisy images, respectively, after downsampling. $F_{Global}^{P}$ and $F_{Global}^{N}$ represent the global features derived from the clean and noisy images, respectively. We calculate the MSE loss ($\mathcal{L}_{MSE}$) between $F_{Global}^{P}$ and $F_{Global}^{N}$, as described in the formula below:
\begin{equation}
    \mathcal{L}_{Noise}=\mathcal{L}_{MSE}(F_{Global}^{P},F_{Global}^{N}) = \left \|  F_{Global}^{P}-F_{Global}^{N}    \right \|^2  
\end{equation}
The total loss function of the model consists of the Binary Cross Entropy loss ($\mathcal{L}_{BCE}$) and the Noise-guided Representation  loss ($\mathcal{L}_{Noise}$). The specific formula is as follows:
\begin{equation}
    \mathcal{L}_{total}= \mathcal{L}_{BCE}+ \mathcal{L}_{Noise},
\end{equation}
we encourage the model to learn a mapping relationship by minimizing the feature space distance between $F^{P}_{Global}$ and $F^{N}_{Global}$. This feature consistency helps constrain the model to learn noise-insensitive essential features. 
\begin{equation}
\begin{split}
\theta' = \text{argmin}_{\theta'} \frac{1}{N_p} \sum_{i=1}^{N_p} &\left( \mathcal{L}_{BCE}(f(x^p_u; \theta'), y^p_u) \right. \\
&\quad + \left. \mathcal{L}_{MSE}(F_{Global}^{P}, F_{Global}^{N}) \right)
\end{split}
\end{equation}
Where $\theta^{'}$ represents the final model parameters. As shown in Fig.\ref{fig1}, introducing noise causes the boundary between the target and background to become more blurred. This blurring effect aids the model in learning the complex relationship between the target edge and the background, encouraging it to focus more on the edge regions of the target, significantly improving edge segmentation performance. Fig.\ref{fig6} provides a visual comparison, showing that our method excels in predicting the target edge regions. The predicted results are much closer to the ground truth (GT).

\section{RealScene-ISTD Benchmark}
\begin{figure*}[!t]
\centering
\includegraphics[width=0.95\linewidth]{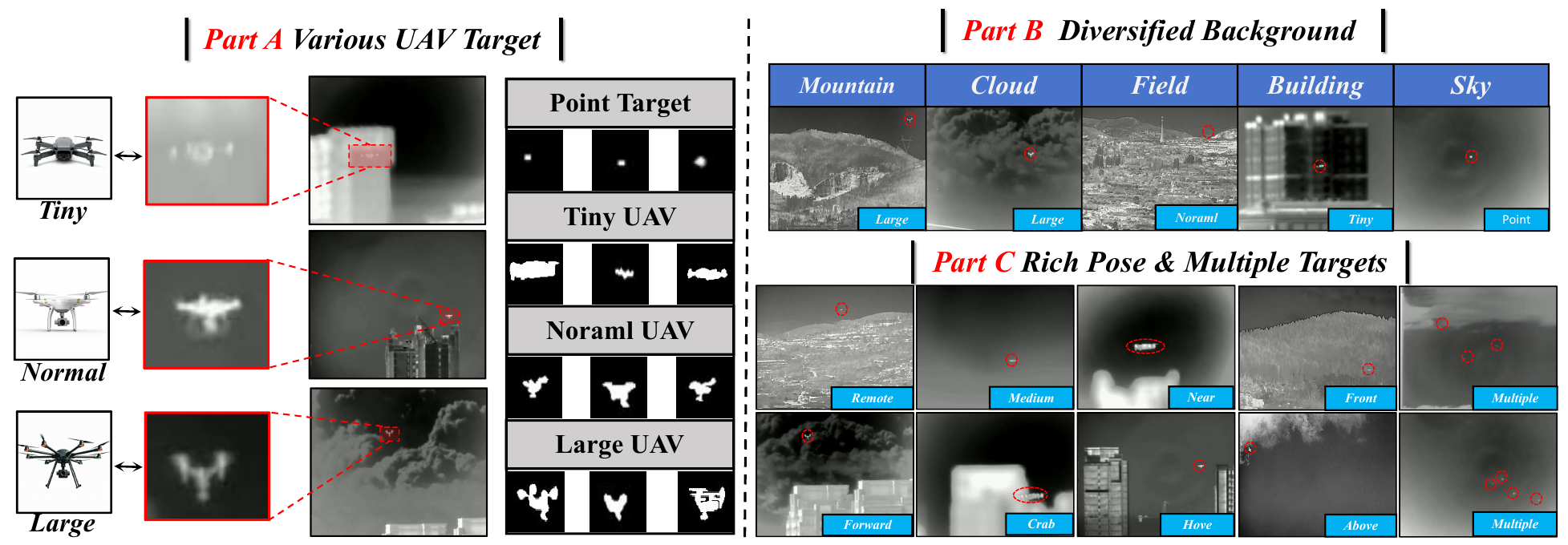}
\caption{Representative samples from our RealScene-ISTD. Part A showcases various UAV target types, while Parts B and C highlight the dataset’s diversity, featuring complex infrared backgrounds, a variety of target types, diverse postures, multi-target and multi-angle scenes, and varying motion trajectories.}
\label{fig7}
\end{figure*}

We collected 739 high-quality infrared UAV images from various real-world scenarios on public platforms, thus constructing a realistic and diverse infrared small target dataset—RealScene-ISTD. All images in the dataset have been manually and accurately cropped and pixel-level annotated, ensuring the accuracy and consistency of the annotations. The image size is standardized to 540 $\times$ 420. As shown in Fig.~\ref{fig7}, the RealScene-ISTD dataset includes UAV targets of three different scales: Tiny, Normal, and Large. These targets are captured by infrared cameras from various angles and distances, resulting in a rich variety of viewpoint changes. The targets are set against complex backgrounds, influenced by atmospheric interference, with significant noise and clutter present. In addition, the dataset includes multiple motion states of the targets, such as stationary, constant-speed movement, and variable-speed movement, which increases the difficulty of detecting the target edges.
To tackle the cross-dataset challenge, we combine the NUAA-SIRST~\cite{NUAA-SIRST}, IRSTD-1K~\cite{IRSTD-1K}, and 739 high-quality annotated UAV images for Real-Scene training and validate them separately. The RealScene-ISTD Benchmark is designed to provide a comprehensive evaluation of algorithm performance and robustness in real-world scenarios, where the models are exposed to diverse data distributions.

\section{EXPERIMENT}
\subsection{Evaluation Metrics}
To comprehensively evaluate the performance of the method proposed in this paper, we adopt a series of standard evaluation metrics. Specifically, we use Intersection over Union ($IoU$) to assess the ability of the model to describe the target shape, and utilize detection accuracy ($P_d$) and false alarm rate ($F_a$) to evaluate the localization accuracy of the model. $IoU$ effectively reflects the overlap between the predicted bounding box and the ground truth bounding box, while $P_d$ and $F_a$ measure the proportion of correctly detected targets and the proportion of incorrectly detected background, respectively. By comparing with the current state-of-the-art (SOTA) methods, we validate the superiority of the proposed method in shape description and target localization. To facilitate a clear visualization of performance variations, optimal results are highlighted in red, sub-optimal results in blue. Performance enhancements are denoted by an upward arrow ($\color{green}{\uparrow}$), while performance degradations are indicated by a downward arrow ($\color{red}{\downarrow}$).

\subsection{Experiment Settings}
\textbf{Cross-dataset Settings:} To simulate real-world scenarios with varying data distributions, we merge the NUAA-SIRST, IRSTD-1K, and RealScene-ISTD datasets for training and perform independent validation. The experiment aims to study the  generalization performance of the model across multiple datasets and enhance the optimization of existing ISTD algorithms in complex environments. Through the multi-dataset fusion approach presented in this paper, we can more comprehensively evaluate the  generalization ability and robustness of the model when confronted with data from different sources and with varying characteristics.

\textbf{Implementation Details:} In this paper, we select the U-Net model with multi-layer attention feature fusion as the baseline, and use ResNets~\cite{resnet} as its backbone network. We set the number of down-sampling layers to 5. The model is optimized using the Binary CrossEntropy loss function and accelerated with the Adam~\cite{adam} optimizer. We also initialized the weights and biases of model using the Kaiming~\cite{he2015delving} initialization method. Additionally, we set the segmentation threshold, initial learning rate, batch size, and number of epochs to 0.5, 0.001, 8, and 1000, respectively. All experiments were conducted on a computer equipped with an NVIDIA TitanXp GPU.

\begin{figure*}[!t]
\centering
\includegraphics[width=0.95\linewidth]{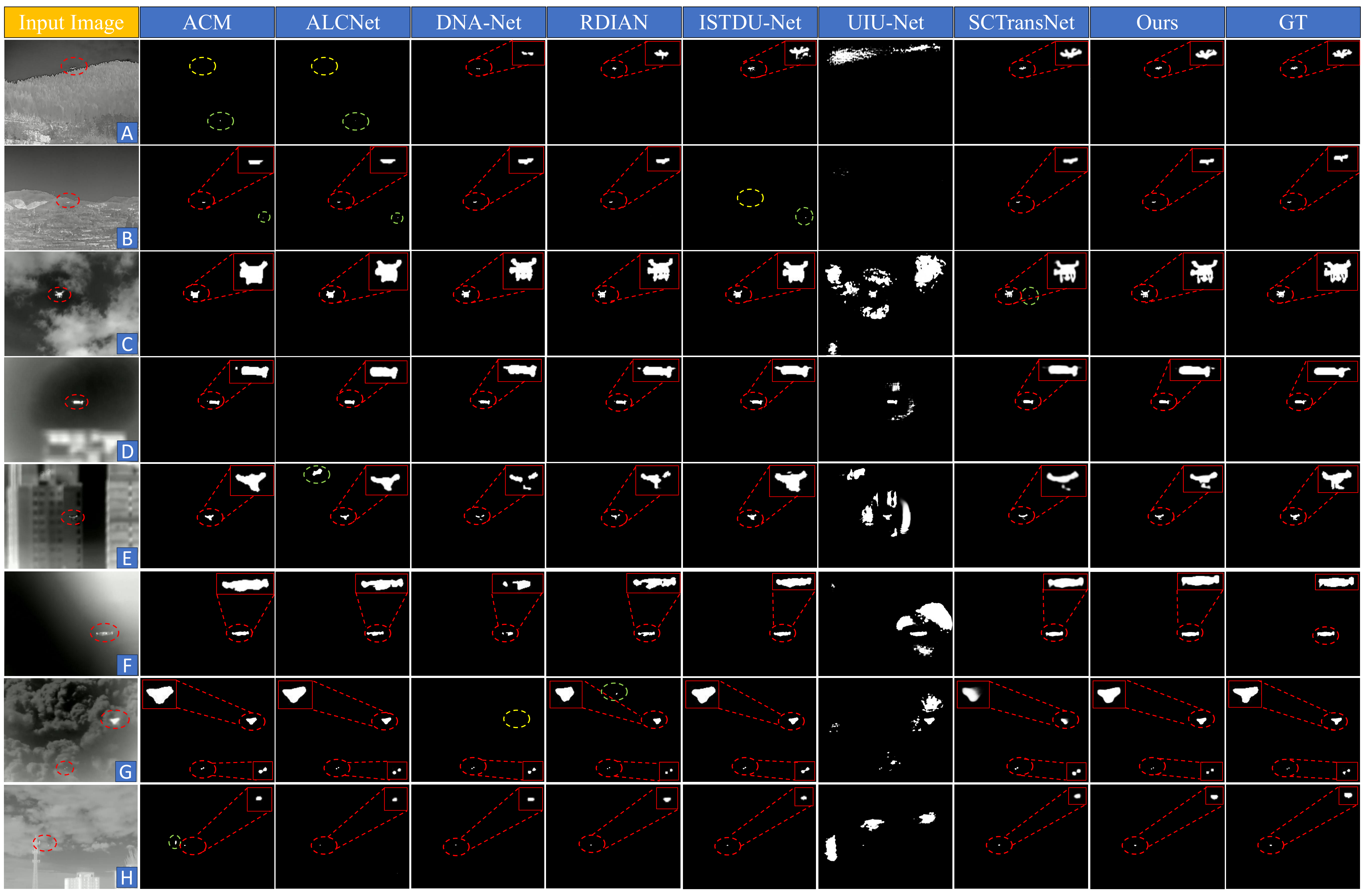}
\caption{The visualization results of each model on the RealScene-ISTD dataset are presented. \textcolor{red}{Red dashed} lines indicate small target regions, \textcolor{green}{green dashed} lines highlight false positives, and \textcolor[rgb]{0.75, 0.5, 0.25}{yellow dashed} lines mark missed detections. To enhance clarity, zoomed-in views are provided. The results show that our method produces outputs that more accurately align with the ground truth.}
\label{fig8}
\end{figure*}

\begin{table}[ht]
    \caption{The performance of different methods under multi-dataset training scenarios is evaluated on the RealScene-ISTD datasets. }
    \label{table9}
    \centering
    \scriptsize
    \resizebox{1\columnwidth}{!}
    {
    \begin{tabular}{cccc}
    \hline
    \multirow{3}{*}{Method} & \multicolumn{3}{c}{Train on NUAA-SIRST,IRSTD-1K,RealScene-ISTD} \\ \cline{2-4} 
                        & \multicolumn{3}{c}{Test on RealScene-ISTD }                                     \\ \cline{2-4}  
                         & $IoU\uparrow(\times 10^{-2})$                                 & $P_d\uparrow(\times 10^{-2})$                                   & $F_a\downarrow(\times 10^{-6})$                \\ \hline
    ACM~\cite{ACM}                  & 64.25               & 91.97              & 250.79               \\
    ALCNet~\cite{ALC}                & 68.88               & 90.91              & 146.85                \\
    DNA-Net~\cite{DNA}                 & 73.79               & 91.33              & 60.06                \\
    RDIAN~\cite{RDIAN}                   & 59.46               & 90.70               & 357.92               \\
    ISTDU-Net~\cite{ISTDU-Net}   & 74.40                & 93.87              & 75.35                \\
    UIU-Net~\cite{UIU}                 & 31.75               & 89.43              & 2928.64               \\
    SCTransNet~\cite{sctransnet}  & {\color{blue}{\textbf{75.01}}}  & \color{blue}{\textbf{94.29}}                                 & \color{blue}{\textbf{53.59}}              \\ \hline  \hline
    Ours                     & {\color{red}{ \textbf{79.32}}} & {\color{red}{\textbf{96.83}}}  & {\color{red}{\textbf{5.40}}}              \\ \hline
    \end{tabular}
    }
    \label{tab9}
\end{table}

\begin{table}[ht]
    \caption{The performance of different methods under multi-dataset training scenarios is evaluated on the IRSTD-1K datasets. }
    \label{table2}
    \centering
    \scriptsize
    \resizebox{1\columnwidth}{!}
    {
    \begin{tabular}{cccc}
    \hline
    \multirow{3}{*}{Method} & \multicolumn{3}{c}{Train on NUAA-SIRST,IRSTD-1K,RealScene-ISTD} \\ \cline{2-4} 
                        & \multicolumn{3}{c}{Test on IRSTD-1K}                                     \\ \cline{2-4}  
                         & $IoU\uparrow(\times 10^{-2})$                                 & $P_d\uparrow(\times 10^{-2})$                                   & $F_a\downarrow(\times 10^{-6})$                \\ \hline
    ACM~\cite{ACM}                 & 57.68               & 92.23              & 115.28               \\
    ALCNet~\cite{ALC}                 & 62.69               & 92.28              & 58.99                \\
    DNA-Net~\cite{DNA}                 & 64.14               &{\color{blue} \textbf{95.64}}              & 40.90                 \\
    RDIAN~\cite{RDIAN}                   & 60.12               & 93.96              & 98.88                \\
    ISTDU-Net~\cite{ISTDU-Net}               & 63.20  & {\color{red}{ \textbf{96.98}}} & {\color{blue}{\textbf{14.36}}}                \\
    UIU-Net~\cite{UIU}                 & 39.10              & 92.62              & 7008.86              \\
    SCTransNet~\cite{sctransnet}              &{\color{blue}{\textbf{66.93}}}  & 93.96 & 15.45                \\ \hline  \hline
    Ours                     & {\color{red}{ \textbf{72.44}}} & 95.30                                 & {\color{red}{ \textbf{14.08}}}                \\ \hline
    \end{tabular}
    }
    \label{tab2}
\end{table}

\subsection{Comparation}
To demonstrate the effectiveness of our method, we compared it with several state-of-the-art approaches on the three public datasets mentioned earlier, including ACM~\cite{ACM}, ALCNet~\cite{ALC}, RDIAN~\cite{RDIAN}, DNA-Net~\cite{DNA}, ISTDU-Net~\cite{ISTDU-Net}, UIU-Net~\cite{UIU}, and SCTransNet~\cite{sctransnet}. For a fair comparison, we re-train all methods on the three public datasets (NUAA-SIRST~\cite{NUAA-SIRST}, IRSTD-1K~\cite{IRSTD-1K}, RealScene-ISTD) using the same implementation details as those in this paper.

\begin{figure}[!t]
\centering
\includegraphics[width=0.95\columnwidth]{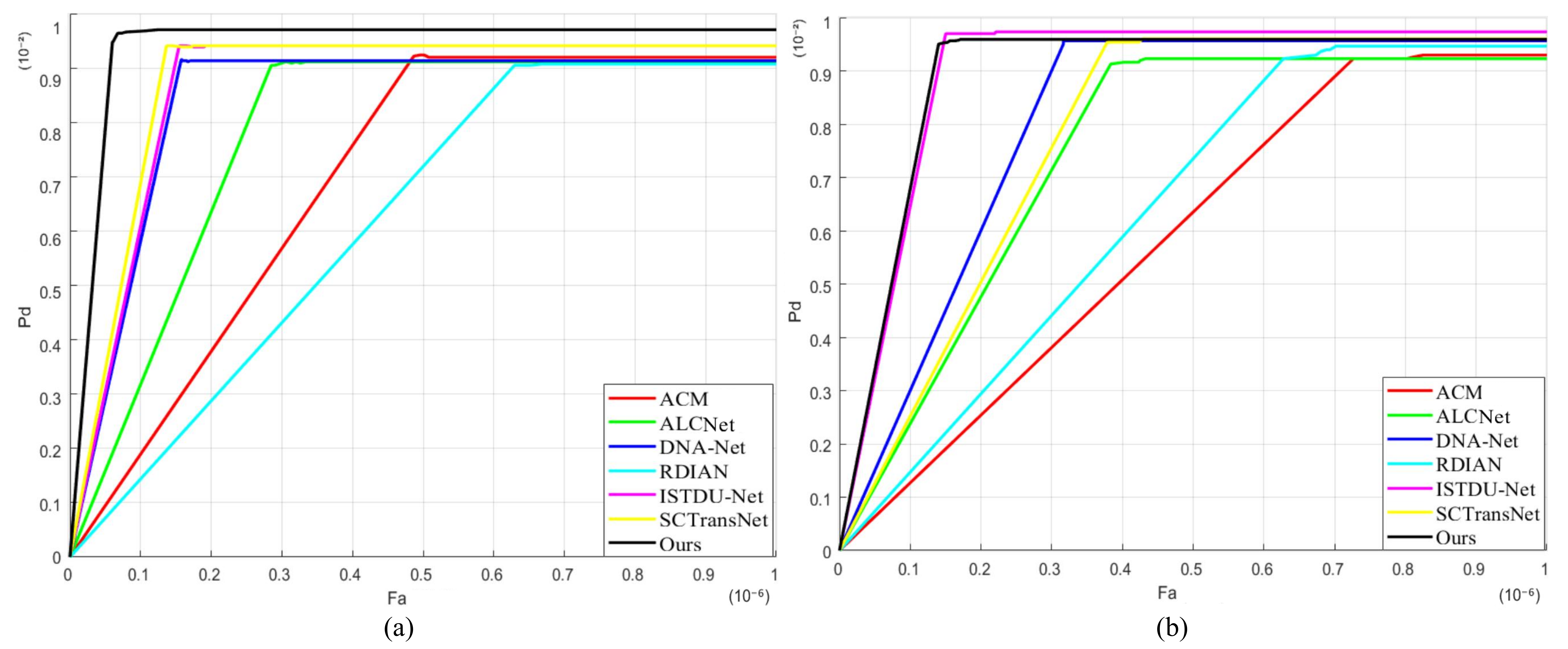}
\caption{Comparison of ROC curves for different methods on the RealScene-ISTD and IRSTD-1K datasets. (a) Results on RealScene-ISTD. (b) Results on IRSTD-1K.}
\label{fig9}
\end{figure}

\textbf{Quantitative comparison.} To comprehensively evaluate performance of the model, we compare our method with others in the multi-dataset joint training scenario and conducted a detailed comparison and analysis of the test results. As shown in Tab.\ref{table2} and Tab.\ref{table9}, the method proposed in this paper outperforms current state-of-the-art detection networks, achieving excellent performance on the real datasets IRSTD-1K\cite{IRSTD-1K} and RealScene-ISTD. Specifically, compared to existing state-of-the-art methods, our method achieves significant improvements in metrics such as $IoU$, detection probability $P_d$, and false alarm rate $F_a$, demonstrating the reliability of the proposed method. Additionally, Fig.\ref{fig9} presents the ROC curves for several state-of-the-art methods on the IRSTD-1K\cite{IRSTD-1K} and RealScene-ISTD datasets. The ROC curve of the proposed method significantly outperforms all other methods, indicating that, compared to existing techniques, our method achieves superior performance in balancing false alarm rates and detection accuracy.

\textbf{Qualitative comparison.} We perform a visual comparison of our method with the latest techniques on RealScene-ISTD. As shown in Fig.\ref{fig8}, considering the relatively limited number of target pixels in the image, we specifically zoom in on local areas to observe the details more clearly. The ACM\cite{ACM}, ALCNet~\cite{ALC}, and ISTDU-Net~\cite{ISTDU-Net} models experience a significant number of missed and false detections. UIU-Net~\cite{UIU} shows limited generalization ability when trained on multiple datasets, resulting in extensive false detections. Although SCTransNet~\cite{sctransnet} reduces errors to some extent, it still struggles to accurately capture the precise contours of targets in certain complex samples. In contrast, our method demonstrates superior performance in target shape reconstruction and localization accuracy, exhibiting higher precision and robustness.

\begin{table}[t]
    \caption{Ablation Experiment of Hyperparameter $\alpha$ in $\mathcal{L}_{Noise}$.  }
    \label{table4}
    \centering
    \scriptsize
    \resizebox{0.85\columnwidth}{!}
    {
    \begin{tabular}{cccc}
    \hline
    Method  & $IoU\uparrow(\times 10^{-2})$  & $P_d\uparrow(\times 10^{-2})$                                   & $F_a\downarrow(\times 10^{-6})$                     \\ \hline
    Baseline    & 76.56   &  93.66 & 47.19   \\ 
    $\alpha$ = 0.2  &  76.04 &  93.65 & \textcolor{red}{\textbf{29.32}} \\ 
    $\alpha$ = 0.4     & 76.57    & 93.87     & 48.82 \\ 
    $\alpha$ = 0.6    & \textcolor{red}{\textbf{77.69}}   & \textcolor{red}{\textbf{95.56}} & \textcolor{blue}{\textbf{29.61}}     \\ 
    $\alpha$ = 0.8   &  \textcolor{blue}{\textbf{76.62}} & 93.66   & 47.09      \\ 
    $\alpha$ = 1   & 76.60    &  93.45 & 37.72   \\ \hline
    \end{tabular}
    }
    \label{tab4}
\end{table}

\begin{table}[t]
    \caption{Comparison of performance on a single dataset with and without $\mathcal{L}_{Noise}$. }
    \label{table5}
    \centering
    \scriptsize
    \resizebox{1\columnwidth}{!}
    {
    \begin{tabular}{cccc}
    \hline
    DataSet                         & Performance & w/o $\mathcal{L}_{Noise}$  & w/ $\mathcal{L}_{Noise}$     \\ \hline
    \multirow{3}{*}{RealScene-ISTD} & $IoU\uparrow$        & 76.56 & \textbf{77.69} \color{green}{($\uparrow$1.13\%) }  \\  
                                &$P_d\uparrow$         & 93.66 & \textbf{95.56} \color{green}{($\uparrow$1.9\%)}    \\ 
                                & $F_a\downarrow$         & 47.19 & \textbf{29.61} \color{green}{($\uparrow$17.58\%)} \\ \hline
    \multirow{3}{*}{NUAA-SIRST~\cite{NUAA-SIRST}}     &$IoU\uparrow$        & 76.40 & \textbf{79.16}\color{green}{($\uparrow$2.76\%)}    \\  
                                & $P_d\uparrow$         & 96.58 & \textbf{96.58}\color{green}{($\uparrow$0.00\%)}    \\  
                                & $F_a\downarrow$         & 28.88 & \textbf{14.34} \color{green}{($\uparrow$14.54\%)}  \\ \hline
    \multirow{3}{*}{IRSTD-1K~\cite{IRSTD-1K}}       & $IoU\uparrow$        & 64.11 & \textbf{66.90} \color{green}{($\uparrow$2.79\%)}   \\  
                                & $P_d\uparrow$         & 94.97 & \textbf{94.97}\color{green}{($\uparrow$0.00\%)}    \\ 
                                & $F_a\downarrow$         & 16.68 & \textbf{26.65} \color{green}{($\uparrow$9.97\%)}   \\ \hline
    \end{tabular}
    }
    \label{tab5}
\end{table}

\begin{table}[t]
    \caption{Quantitative comparison of CCA results on RealScene-ISTD and IRSTD-1k.}
    \label{table6}
    \centering
    \scriptsize
    \resizebox{1\columnwidth}{!}
    {
    \begin{tabular}{cccccc}
    \hline
    \multicolumn{1}{l}{TrainSet} & TestSet                         & Performance & Baseline & w/o CCA      & w/ CCA        \\ \hline
                             & \multirow{3}{*}{RealScene-ISTD} & $IoU \uparrow$        & 76.56    & 75.01\color{red}{($\downarrow$1.55\%)} & \textbf{77.30} \color{green}{($\uparrow$0.74\%)} \\
    Train on                     &                                 & $P_d \uparrow$         & 93.66    & 94.29        & \textbf{96.41} \color{green}{($\uparrow$2.75\%)} \\
    NUAA-SIRST,                  &                                 & $F_a \downarrow$         & 47.19    & 53.59\color{red}{($\downarrow$6.4\%)}  & \textbf{42.27} \color{green}{($\uparrow$4.92\%)} \\ \cline{2-6} 
    IRSTD-1K,                    & \multirow{3}{*}{IRSTD-1k~\cite{IRSTD-1K}}       & $IoU \uparrow$        & 64.11    & 66.93        & \textbf{66.75}\color{green}{($\uparrow$2.64\%) } \\
    RealScene-ISTD               &                                 & $P_d \uparrow$         & 94.97    & 93.96\color{red}{($\downarrow$1.01\%)} & \textbf{95.97} \color{green}{($\uparrow$1.00\%)} \\
                             &                                 & $F_a\downarrow$         & 16.68    & 15.45        & \textbf{13.38} \color{green}{($\uparrow$3.30\%)} \\ \hline
    \end{tabular}
    }
    \label{tab6}
\end{table}

\begin{table}[t]
    \caption{Ablation study of CCA, Cross-view Top-K Fusion, and $\mathcal{L}_{Noise}$ modules on the RealScene-ISTD dataset.}
    \label{table7}
    \centering
    \scriptsize
    \resizebox{1\columnwidth}{!}
    {
    \begin{tabular}{ccccccc}
    \hline
    \multicolumn{1}{l}{\multirow{2}{*}{baseline}} & \multirow{2}{*}{CCA} & \multirow{2}{*}{Top-K Fusion} & \multirow{2}{*}{$\mathcal{L}_{Noise}$} & \multicolumn{3}{c}{Performance}             \\ \cline{5-7} 
    \multicolumn{1}{l}{}                          &                      &                        &                        & $IoU\uparrow$         & $P_d \uparrow$          & $F_a \downarrow$           \\ \hline
    $\checkmark$                                             &                     &                      &                     & 75.01        & 94.29        & 53.59         \\
    $\checkmark$                                             & $\checkmark$                    &                     &                      & 77.30        & 96.41        & 42.27         \\
    $\checkmark$                                             & $\checkmark$                    & $\checkmark$                                     &      & 79.46        & 96.61        & 27.84         \\
    $\checkmark$                                             & $\checkmark$                    & $\checkmark$                      & $\checkmark$                      & \textbf{79.32}\color{green}{($\uparrow$4.31\%)} & \textbf{96.83}\color{green}{($\uparrow$2.54\%)} & \textbf{5.40}\color{green}{($\uparrow$48.19\%)} \\ \hline
    \end{tabular}
    }
    \label{tab7}
\end{table}

\begin{table}[t]
    \caption{Quantitative comparison of CCA+Top-K Fusion results on RealScene-ISTD.}
    \label{table8}
    \centering
    \scriptsize
    \resizebox{1\columnwidth}{!}
    {
    \begin{tabular}{lccc}
    \hline
                    & \multicolumn{3}{c}{Performance}      \\ \cline{2-4} 
    \multirow{-2}{*}{Method}           & $IoU\uparrow$                              & $P_d \uparrow$                                & $F_a \downarrow$                                 \\ \hline
    w/o CCA+Top-K Fusion & 74.40                             & 93.87      & 75.35                               \\
    w/ CCA+Top-K Fusion  & \multicolumn{1}{l}{\textbf{76.90}\color{green}{($\uparrow2.5\%$)}} & \multicolumn{1}{l}{\textbf{94.50}\color{green}{($\uparrow0.63\%$)}} & \multicolumn{1}{l}{\textbf{39.69}\color{green}{($\uparrow35.66\%$)}} \\ \hline
    \end{tabular}
    }
    \label{tab8}
\end{table}

\begin{table}[t]
    \caption{Efficiency Analysis of Different SOTA ISTD Methods on the RealScene-ISTD. }
    \label{table3}
    \centering
    \scriptsize
    \resizebox{1\columnwidth}{!}
    {
    \begin{tabular}{ccccc}
    \hline
    Method   &  Pub year& Params (M) & Inference times(s)  &   $IoU$ ($\times 10^{-2}$)              \\ \hline
    ACM~\cite{ACM}    & 2021   &  0.398 & 0.0590 & 66.26 \\ 
    ALCNet~\cite{ALC}  &  2021 &   0.427&  0.0588 & 54.51\\ \
    DNA-Net~\cite{DNA}     & 2022    & 4.697     &0.1707 &75.02 \\ 
    ISTDU-Net~\cite{ISTDU-Net}    & 2022     & 2.752 & 0.0764 &63.37    \\ 
    UIU-Net~\cite{UIU}   &  2022 & 50.54   & 0.0905  & 68.17   \\ 
    RDIAN~\cite{RDIAN}   & 2023    &  0.217 & 0.0694  & 70.03 \\ 
    SCTransNet~\cite{sctransnet}   & 2024    &  11.19 & 0.1066  & 76.56 \\ \hline \hline
    Ours   & 2025    &  22.38 & 0.1053  & 77.69 \\ \hline 
    \end{tabular}
    }
    \label{tab3}
\end{table}

\subsection{Ablation Study}
In this section, we systematically assess the effectiveness of each component: Cross-view Channel Alignment (CCA), Cross-view Top-K Fusion, and Noise-guided Representation  Loss ($\mathcal{L}_{Noise}$). Additionally, we perform an ablation study on the hyperparameter $\alpha$ of the Noise-guided Representation  Loss component to evaluate its performance independently under the best configuration. To facilitate a clear visualization of performance variations, optimal results are highlighted in red, sub-optimal results in blue. Performance enhancements are denoted by an upward arrow ($\color{green}{\uparrow}$), while performance degradations are indicated by a downward arrow ($\color{red}{\downarrow}$).

\textbf{Ablation analysis of the thermal noise factor hyperparameter $\alpha$ in  $\mathcal{L}_{Noise}$.} To identify the optimal thermal noise factor $\alpha$ in $\mathcal{L}_{Noise}$, we conduct an ablation study on $\alpha$ using the baseline model, specifically on the RealScene-ISTD dataset. As shown in Tab.~\ref{table4}, the model achieves the best performance when $\alpha = 0.6$. Compared to the baseline, the model incorporating $\mathcal{L}_{Noise}$ exhibits significant improvements across all evaluation metrics. The experimental results reveal that a low $\alpha$ fails to introduce enough training pressure, restricting the ability to capture intrinsic data features. On the other hand, an excessively high $\alpha$ makes the model overly dependent on noise, diverting attention from the true data distribution and ultimately degrading performance. The incorporation of $\mathcal{L}_{Noise}$ enables the model to leverage noise more effectively, strengthening its robustness and adaptability in challenging noisy conditions.

\textbf{The Effect of the $\mathcal{L}_{Noise}$.} To further validate the effectiveness of $\mathcal{L}_{Noise}$, we conduct a comparative analysis of detection performance between the enhanced model and the baseline(‘w/o $\mathcal{L}_{Noise}$’ vs. ‘w/ $\mathcal{L}_{Noise}$’) across three distinct datasets. As shown in Tab.~\ref{table5}, the model incorporating high-resolution representation learning achieved a significant improvement in detection performance, particularly excelling in the $F_a$ metric.

\textbf{The Effect of the CCA.} We first compare the performance differences between models trained with direct dataset fusion and those utilizing the CCA module for dataset alignment (‘w/o CCA’ vs. ‘w/ CCA’). As shown in Tab.~\ref{table6}, simply merging datasets for training does not effectively improve the model’s detection performance. In contrast, using the CCA module can narrow the feature distribution gap between different datasets, which can better improve the  performance of model. This indicates that the module effectively solves the data heterogeneity problem during multi-dataset fusion, improving the  generalization performance of model.

\textbf{Ablation on CCA, Cross-view Top-K Fusion, and $\mathcal{L}_{Noise}$.} To assess the effectiveness of CCA, Cross-view Top-K Fusion, and $\mathcal{L}_{Noise}$, we conduct an ablation study by analyzing each component separately. The experiments were performed on the RealScene-ISTD test set, and the ablation results are shown in Tab.~\ref{table7}.

\textbf{Effectiveness on More Backbones.} To further assess the effectiveness of the Cross-view Dataset Alignment and Fusion strategy, we evaluate its impact across different backbones. In this section, ISTDU-Net~\cite{ISTDU-Net} is adopted as the detection model and compared its performance with and without Cross-view Dataset Alignment and Fusion (‘w/o CCA+Top-K Fusion’ vs. ‘w/ CCA+Top-K Fusion’). As shown in Tab~\ref{table8}, we evaluated ISTDU-Net~\cite{ISTDU-Net} on the RealScene-ISTD dataset. The integration of Cross-view Dataset Alignment and Fusion demonstrably yields substantial improvements across all evaluation metrics for the ISTDU-Net~\cite{ISTDU-Net}.

\textbf{Efficiency analysis.} We conduct an efficiency analysis of recent ISTD models, as shown in Tab.\ref{table3}. Compared to other methods, our model demonstrates substantial advantages in both performance and efficiency. Specifically, our model outperforms the current top model, SCTransNet\cite{sctransnet}, by a considerable margin in terms of the IoU metric, achieving the best overall performance. Notably, our model significantly reduces the number of parameters compared to the UIU-Net~\cite{UIU} model, showcasing superior computational efficiency. Although our model has a slight increase in parameters compared to lightweight models like ACM~\cite{ACM} and ALCNet~\cite{ALC}, its performance improvement is significant, far exceeding these two models in results and fully reflecting an excellent balance between performance and efficiency.
\section{CONCLUSION }
To address the domain discrepancy issue in ISTD models, this paper introduces a domain-adaptive representation learning framework. Specifically, the Cross-view Dataset Alignment and Fusion strategy is proposed to align and integrate datasets from different perspectives, reducing domain bias. Meanwhile, the Noise-guided Representation Learning strategy is designed to facilitate the model to capture noise-resistant features, enhancing its generalization in sensor-specific noise domains. Extensive experimental results demonstrate that the proposed method effectively resolves domain distribution discrepancies across datasets and exhibits superior generalization performance in diverse environments. In conclusion, the proposed approach significantly mitigates the domain shift challenge in ISTD, paving the way for its practical application in real-world scenarios.
\bibliographystyle{IEEEtran}
\bibliography{ref}

\begin{thebibliography}{10}
\providecommand{\url}[1]{#1}
\csname url@samestyle\endcsname
\providecommand{\newblock}{\relax}
\providecommand{\bibinfo}[2]{#2}
\providecommand{\BIBentrySTDinterwordspacing}{\spaceskip=0pt\relax}
\providecommand{\BIBentryALTinterwordstretchfactor}{4}
\providecommand{\BIBentryALTinterwordspacing}{\spaceskip=\fontdimen2\font plus
\BIBentryALTinterwordstretchfactor\fontdimen3\font minus
  \fontdimen4\font\relax}
\providecommand{\BIBforeignlanguage}[2]{{%
\expandafter\ifx\csname l@#1\endcsname\relax
\typeout{** WARNING: IEEEtran.bst: No hyphenation pattern has been}%
\typeout{** loaded for the language `#1'. Using the pattern for}%
\typeout{** the default language instead.}%
\else
\language=\csname l@#1\endcsname
\fi
#2}}
\providecommand{\BIBdecl}{\relax}
\BIBdecl

\bibitem{security1}
Y.~Sun, S.~Abeywickrama, L.~Jayasinghe, C.~Yuen, J.~Chen, and M.~Zhang,
  ``Micro-doppler signature-based detection, classification, and localization
  of small uav with long short-term memory neural network,'' \emph{IEEE
  Transactions on Geoscience and Remote Sensing}, vol.~59, no.~8, pp.
  6285--6300, 2021.

\bibitem{security2}
H.~Fang, C.~Wu, X.~Wang, F.~Zhou, Y.~Chang, and L.~Yan, ``Online infrared uav
  target tracking with enhanced context-awareness and pixel-wise attention
  modulation,'' \emph{IEEE Transactions on Geoscience and Remote Sensing},
  vol.~62, pp. 1--17, 2024.

\bibitem{rescue}
Y.~Tong, J.~Liu, Z.~Fu, Z.~Wang, H.~Yang, S.~Niu, and Q.~Tan, ``Guided
  attention and joint loss for infrared dim small target detection,''
  \emph{IEEE Transactions on Geoscience and Remote Sensing}, vol.~62, pp.
  1--14, 2024.

\bibitem{autonomous}
Q.~Hu, W.~Yu, S.~Zhu, Y.~Huang, X.~Zhang, M.~Liu, and B.~Cao, ``Uav-based
  thermal radiation directionality capture and its evaluation on kernel-driven
  models,'' \emph{IEEE Transactions on Geoscience and Remote Sensing}, vol.~63,
  pp. 1--11, 2025.

\bibitem{monitoring1}
H.~Yao and X.~Liang, ``Autonomous exploration under canopy for forest
  investigation using lidar and quadrotor,'' \emph{IEEE Transactions on
  Geoscience and Remote Sensing}, vol.~62, pp. 1--19, 2024.

\bibitem{monitoring2}
W.~Lu, Z.~Zhang, and M.~Nguyen, ``A lightweight cnn–transformer network with
  laplacian loss for low-altitude uav imagery semantic segmentation,''
  \emph{IEEE Transactions on Geoscience and Remote Sensing}, vol.~62, pp.
  1--20, 2024.

\bibitem{MITHF}
Y.~Li, Z.~Li, C.~Zhang, Z.~Luo, Y.~Zhu, Z.~Ding, and T.~Qin, ``Infrared
  maritime dim small target detection based on spatiotemporal cues and
  directional morphological filtering,'' \emph{Infrared Physics \& Technology},
  vol. 115, p. 103657, 2021.

\bibitem{Mclntosh}
B.~McIntosh, S.~Venkataramanan, and A.~Mahalanobis, ``Infrared target detection
  in cluttered environments by maximization of a target to clutter ratio (tcr)
  metric using a convolutional neural network,'' \emph{IEEE Transactions on
  Aerospace and Electronic Systems}, vol.~57, no.~1, pp. 485--496, 2020.

\bibitem{WSLCM}
J.~Han, S.~Moradi, I.~Faramarzi, H.~Zhang, Q.~Zhao, X.~Zhang, and N.~Li,
  ``Infrared small target detection based on the weighted strengthened local
  contrast measure,'' \emph{IEEE Geoscience and Remote Sensing Letters},
  vol.~18, no.~9, pp. 1670--1674, 2020.

\bibitem{TLLCM}
J.~Han, S.~Moradi, I.~Faramarzi, C.~Liu, H.~Zhang, and Q.~Zhao, ``A local
  contrast method for infrared small-target detection utilizing a tri-layer
  window,'' \emph{IEEE Geoscience and Remote Sensing Letters}, vol.~17, no.~10,
  pp. 1822--1826, 2019.

\bibitem{NRAM}
L.~Zhang, L.~Peng, T.~Zhang, S.~Cao, and Z.~Peng, ``Infrared small target
  detection via non-convex rank approximation minimization joint l 2, 1 norm,''
  \emph{Remote Sensing}, vol.~10, no.~11, p. 1821, 2018.

\bibitem{deep2}
S.~Liu, B.~Qiao, S.~Li, Y.~Wang, and L.~Dang, ``Patch spatial attention
  networks for semantic token transformer in infrared small target detection,''
  \emph{IEEE Transactions on Geoscience and Remote Sensing}, pp. 1--1, 2025.

\bibitem{deep1}
Q.~Li, W.~Zhang, W.~Lu, and Q.~Wang, ``Multibranch mutual-guiding learning for
  infrared small target detection,'' \emph{IEEE Transactions on Geoscience and
  Remote Sensing}, vol.~63, pp. 1--10, 2025.

\bibitem{deep3}
Z.~Wang, C.~Wang, X.~Li, C.~Xia, and J.~Xu, ``Mlp-net: Multilayer perceptron
  fusion network for infrared small target detection,'' \emph{IEEE Transactions
  on Geoscience and Remote Sensing}, vol.~63, pp. 1--13, 2025.

\bibitem{filtering1}
X.~Zhang, J.~Ru, and C.~Wu, ``Infrared small target detection based on gradient
  correlation filtering and contrast measurement,'' \emph{IEEE Transactions on
  Geoscience and Remote Sensing}, vol.~61, pp. 1--12, 2023.

\bibitem{filtering2}
T.-W. Bae, ``Small target detection using bilateral filter and temporal cross
  product in infrared images,'' \emph{Infrared Physics \& Technology}, vol.~54,
  no.~5, pp. 403--411, 2011.

\bibitem{filtering3}
H.~Zhu, H.~Ni, S.~Liu, G.~Xu, and L.~Deng, ``Tnlrs: Target-aware non-local
  low-rank modeling with saliency filtering regularization for infrared small
  target detection,'' \emph{IEEE Transactions on Image Processing}, vol.~29,
  pp. 9546--9558, 2020.

\bibitem{HVS1}
J.~Han, Y.~Ma, B.~Zhou, F.~Fan, K.~Liang, and Y.~Fang, ``A robust infrared
  small target detection algorithm based on human visual system,'' \emph{IEEE
  Geoscience and Remote Sensing Letters}, vol.~11, no.~12, pp. 2168--2172,
  2014.

\bibitem{HVS2}
S.~Kim, Y.~Yang, J.~Lee, and Y.~Park, ``Small target detection utilizing robust
  methods of the human visual system for irst,'' \emph{Journal of infrared,
  millimeter, and terahertz waves}, vol.~30, pp. 994--1011, 2009.

\bibitem{ALC}
Y.~Dai, Y.~Wu, F.~Zhou, and K.~Barnard, ``Attentional local contrast networks
  for infrared small target detection,'' \emph{IEEE Transactions on Geoscience
  and Remote Sensing}, vol.~59, no.~11, pp. 9813--9824, 2021.

\bibitem{DNA}
B.~Li, C.~Xiao, L.~Wang, Y.~Wang, Z.~Lin, M.~Li, W.~An, and Y.~Guo, ``Dense
  nested attention network for infrared small target detection,'' \emph{IEEE
  Transactions on Image Processing}, vol.~32, pp. 1745--1758, 2022.

\bibitem{vaswani2017attention}
A.~Vaswani, N.~Shazeer, N.~Parmar, J.~Uszkoreit, L.~Jones, A.~N. Gomez,
  {\L}.~Kaiser, and I.~Polosukhin, ``Attention is all you need,''
  \emph{Advances in neural information processing systems}, vol.~30, 2017.

\bibitem{unet}
O.~Ronneberger, P.~Fischer, and T.~Brox, ``U-net: Convolutional networks for
  biomedical image segmentation,'' pp. 234--241, 2015.

\bibitem{ACM}
Y.~Dai, Y.~Wu, F.~Zhou, and K.~Barnard, ``Asymmetric contextual modulation for
  infrared small target detection,'' pp. 950--959, 2021.

\bibitem{RDIAN}
H.~Sun, J.~Bai, F.~Yang, and X.~Bai, ``Receptive-field and direction induced
  attention network for infrared dim small target detection with a large-scale
  dataset irdst,'' \emph{IEEE Transactions on Geoscience and Remote Sensing},
  vol.~61, pp. 1--13, 2023.

\bibitem{UIU}
X.~Wu, D.~Hong, and J.~Chanussot, ``Uiu-net: U-net in u-net for infrared small
  object detection,'' \emph{IEEE Transactions on Image Processing}, vol.~32,
  pp. 364--376, 2022.

\bibitem{sctransnet}
S.~Yuan, H.~Qin, X.~Yan, N.~Akhtar, and A.~Mian, ``Sctransnet: Spatial-channel
  cross transformer network for infrared small target detection,'' \emph{IEEE
  Transactions on Geoscience and Remote Sensing}, 2024.

\bibitem{sky}
J.~Zhao, Z.~Shi, C.~Yu, and Y.~Liu, ``Infrared small target detection based on
  adjustable sensitivity strategy and multi-scale fusion,'' \emph{arXiv
  preprint arXiv:2407.20090}, 2024.

\bibitem{mosaic}
Y.~Shi, Y.~Lin, P.~Wei, X.~Xian, T.~Chen, and L.~Lin, ``Diff-mosaic: augmenting
  realistic representations in infrared small target detection via diffusion
  prior,'' \emph{IEEE Transactions on Geoscience and Remote Sensing}, 2024.

\bibitem{Beyondfulllabels}
S.~Yuan, H.~Qin, R.~Kou, X.~Yan, Z.~Li, C.~Peng, D.~Wu, and H.~Zhou, ``Beyond
  full labels: Energy-double-guided single-point prompt for infrared small
  target label generation,'' \emph{IEEE Journal of Selected Topics in Applied
  Earth Observations and Remote Sensing}, 2025.

\bibitem{Tophat}
J.-F. Rivest and R.~Fortin, ``Detection of dim targets in digital infrared
  imagery by morphological image processing,'' \emph{Optical Engineering},
  vol.~35, no.~7, pp. 1886--1893, 1996.

\bibitem{LCM}
C.~L.~P. Chen, H.~Li, Y.~Wei, T.~Xia, and Y.~Y. Tang, ``A local contrast method
  for small infrared target detection,'' \emph{IEEE Transactions on Geoscience
  and Remote Sensing}, vol.~52, no.~1, pp. 574--581, 2014.

\bibitem{IPI}
C.~Gao, D.~Meng, Y.~Yang, Y.~Wang, X.~Zhou, and A.~G. Hauptmann, ``Infrared
  patch-image model for small target detection in a single image,'' \emph{IEEE
  transactions on image processing}, vol.~22, no.~12, pp. 4996--5009, 2013.

\bibitem{cnn}
Y.~LeCun, L.~Bottou, Y.~Bengio, and P.~Haffner, ``Gradient-based learning
  applied to document recognition,'' \emph{Proceedings of the IEEE}, vol.~86,
  no.~11, pp. 2278--2324, 1998.

\bibitem{MDvsFA-GAN}
H.~Wang, L.~Zhou, and L.~Wang, ``Miss detection vs. false alarm: Adversarial
  learning for small object segmentation in infrared images,'' in
  \emph{Proceedings of the IEEE/CVF international conference on computer
  vision}, 2019, pp. 8509--8518.

\bibitem{DomainAdaptation1}
J.~Hu, J.~Lu, and Y.-P. Tan, ``Deep transfer metric learning,'' pp. 325--333,
  2015.

\bibitem{DomainAdaptation2}
S.~Motiian, M.~Piccirilli, D.~A. Adjeroh, and G.~Doretto, ``Unified deep
  supervised domain adaptation and generalization,'' pp. 5715--5725, 2017.

\bibitem{DomainAdaptation3}
T.~Gebru, J.~Hoffman, and L.~Fei-Fei, ``Fine-grained recognition in the wild: A
  multi-task domain adaptation approach,'' pp. 1349--1358, 2017.

\bibitem{DG1}
G.~Bai, C.~Ling, and L.~Zhao, ``Temporal domain generalization with drift-aware
  dynamic neural networks,'' \emph{arXiv preprint arXiv:2205.10664}, 2022.

\bibitem{DG2}
K.~Zhou, Z.~Liu, Y.~Qiao, T.~Xiang, and C.~C. Loy, ``Domain generalization: A
  survey,'' \emph{IEEE transactions on pattern analysis and machine
  intelligence}, vol.~45, no.~4, pp. 4396--4415, 2022.

\bibitem{DG3}
Z.~Wang, Y.~Luo, R.~Qiu, Z.~Huang, and M.~Baktashmotlagh, ``Learning to
  diversify for single domain generalization,'' in \emph{Proceedings of the
  IEEE/CVF international conference on computer vision}, 2021, pp. 834--843.

\bibitem{TTA1}
D.~Chen, D.~Wang, T.~Darrell, and S.~Ebrahimi, ``Contrastive test-time
  adaptation,'' in \emph{Proceedings of the IEEE/CVF Conference on Computer
  Vision and Pattern Recognition}, 2022, pp. 295--305.

\bibitem{TTA2}
L.~Chen, Y.~Zhang, Y.~Song, Y.~Shan, and L.~Liu, ``Improved test-time
  adaptation for domain generalization,'' in \emph{Proceedings of the IEEE/CVF
  Conference on Computer Vision and Pattern Recognition}, 2023, pp.
  24\,172--24\,182.

\bibitem{TTA3}
S.~Wang, D.~Zhang, Z.~Yan, J.~Zhang, and R.~Li, ``Feature alignment and
  uniformity for test time adaptation,'' in \emph{Proceedings of the IEEE/CVF
  Conference on Computer Vision and Pattern Recognition}, 2023, pp.
  20\,050--20\,060.

\bibitem{TTA4}
Y.~Yuan, B.~Xu, L.~Hou, F.~Sun, H.~Shen, and X.~Cheng, ``Tea: Test-time energy
  adaptation,'' in \emph{Proceedings of the IEEE/CVF Conference on Computer
  Vision and Pattern Recognition}, 2024, pp. 23\,901--23\,911.

\bibitem{ds1}
S.~Sankaranarayanan, Y.~Balaji, A.~Jain, S.~N. Lim, and R.~Chellappa,
  ``Learning from synthetic data: Addressing domain shift for semantic
  segmentation,'' in \emph{Proceedings of the IEEE conference on computer
  vision and pattern recognition}, 2018, pp. 3752--3761.

\bibitem{ds2}
J.~Zhang, K.~Yang, H.~Shi, S.~Rei{\ss}, K.~Peng, C.~Ma, H.~Fu, P.~H. Torr,
  K.~Wang, and R.~Stiefelhagen, ``Behind every domain there is a shift:
  Adapting distortion-aware vision transformers for panoramic semantic
  segmentation,'' \emph{IEEE Transactions on Pattern Analysis and Machine
  Intelligence}, 2024.

\bibitem{fusion}
J.~Ma, L.~Tang, F.~Fan, J.~Huang, X.~Mei, and Y.~Ma, ``Swinfusion: Cross-domain
  long-range learning for general image fusion via swin transformer,''
  \emph{IEEE/CAA Journal of Automatica Sinica}, vol.~9, no.~7, pp. 1200--1217,
  2022.

\bibitem{NUAA-SIRST}
Y.~Dai, Y.~Wu, F.~Zhou, and K.~Barnard, ``Asymmetric contextual modulation for
  infrared small target detection,'' pp. 950--959, 2021.

\bibitem{IRSTD-1K}
M.~Zhang, R.~Zhang, Y.~Yang, H.~Bai, J.~Zhang, and J.~Guo, ``Isnet: Shape
  matters for infrared small target detection,'' pp. 877--886, 2022.

\bibitem{resnet}
K.~He, X.~Zhang, S.~Ren, and J.~Sun, ``Deep residual learning for image
  recognition,'' \emph{2016 IEEE Conference on Computer Vision and Pattern
  Recognition (CVPR)}, pp. 770--778, 2015.

\bibitem{adam}
D.~P. Kingma and J.~Ba, ``Adam: A method for stochastic optimization,''
  \emph{arXiv preprint arXiv:1412.6980}, 2014.

\bibitem{he2015delving}
K.~He, X.~Zhang, S.~Ren, and J.~Sun, ``Delving deep into rectifiers: Surpassing
  human-level performance on imagenet classification,'' pp. 1026--1034, 2015.

\bibitem{ISTDU-Net}
Q.~Hou, L.~Zhang, F.~Tan, Y.~Xi, H.~Zheng, and N.~Li, ``Istdu-net: Infrared
  small-target detection u-net,'' \emph{IEEE Geoscience and Remote Sensing
  Letters}, vol.~19, pp. 1--5, 2022.

\end{thebibliography}

\begin{IEEEbiography}[{\includegraphics[width=1in,height=1.25in,clip,keepaspectratio]{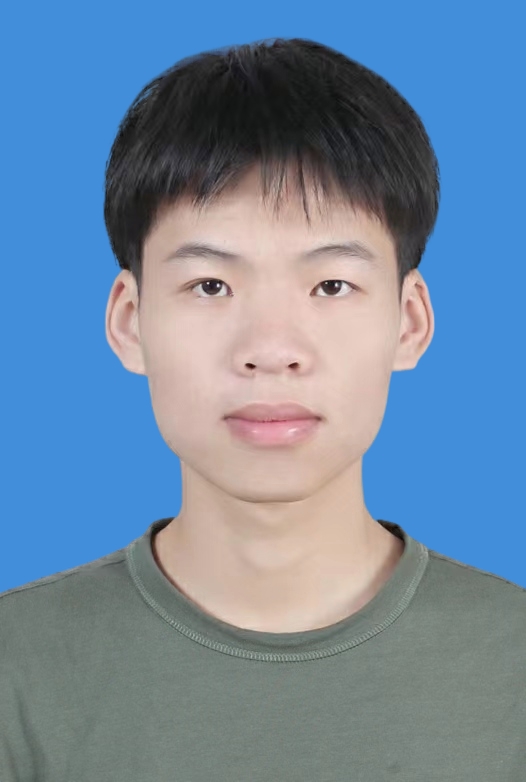}}]{Yahao Lu} received the B.S. degree in 2023, from the School of Information Engineering, Guangdong University of Technology, Guangzhou, China, where he is currently working towards a M.S. degree. His research interests include computer vision and machine learning.
\end{IEEEbiography}

\begin{IEEEbiography}[{\includegraphics[width=1in,height=1.25in,clip,keepaspectratio]{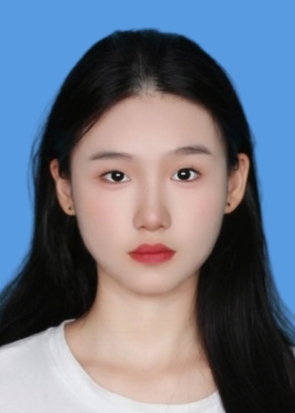}}]{Yuehui Li} is currently pursuing the B.S. degree at the School of Information Engineering, Guangdong University of Technology, Guangzhou, China, with expected graduation in 2026. Her research interests include computer vision and machine learning.
\end{IEEEbiography}

\begin{IEEEbiography}[{\includegraphics[width=1in,height=1.25in,clip,keepaspectratio]{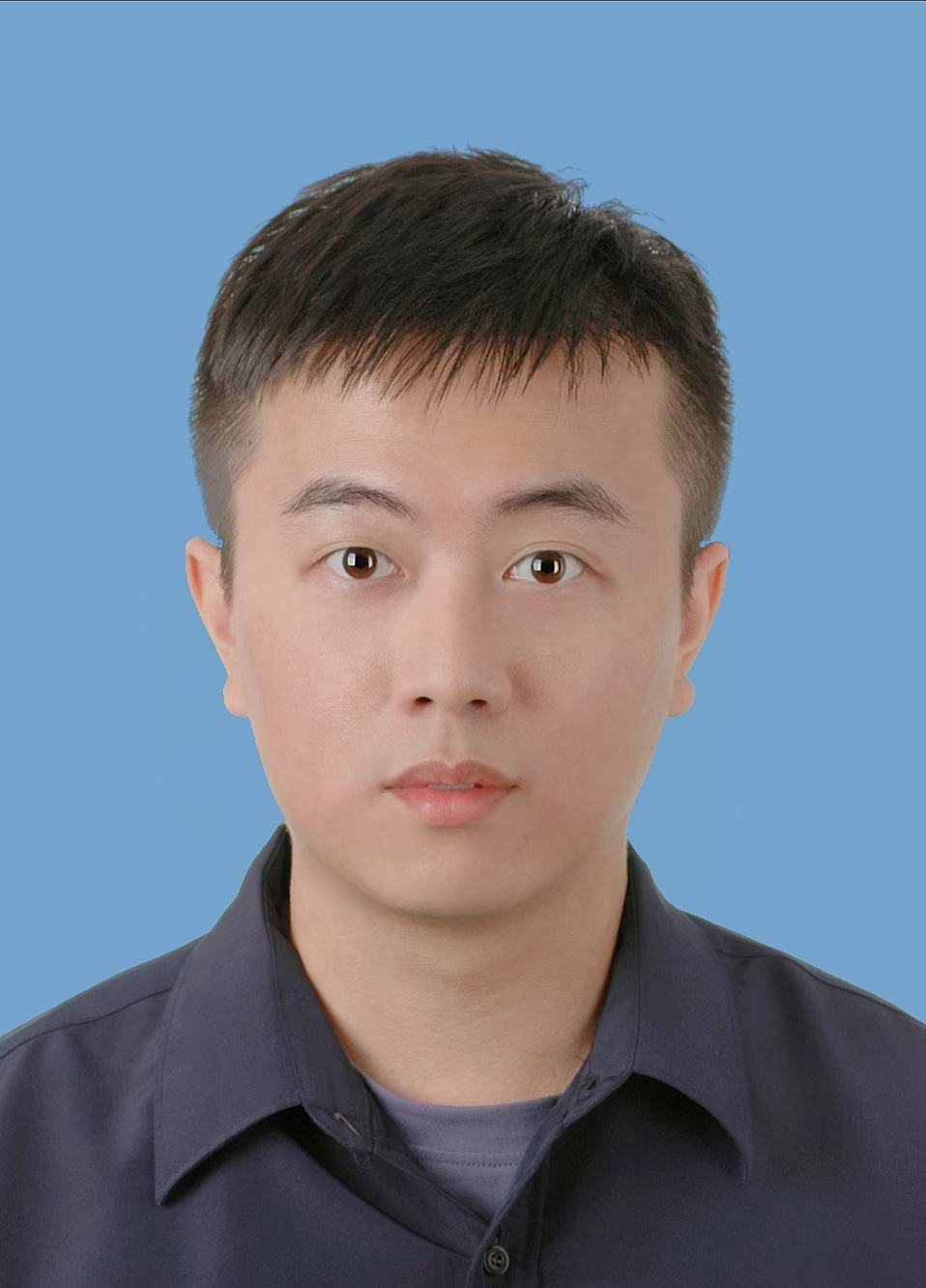}}]{Xingyuan Guo} received the B.S. degree in 2010, from the School of Automation, Guangdong University of Technology, Guangzhou, China. He is currently a senior engineer with the Southern Power Grid Company, Ltd., China. His current research interests include modeling, stability and control of industrial systems.
\end{IEEEbiography}

\begin{IEEEbiography}[{\includegraphics[width=1.1in, height=1.4in, clip, keepaspectratio]{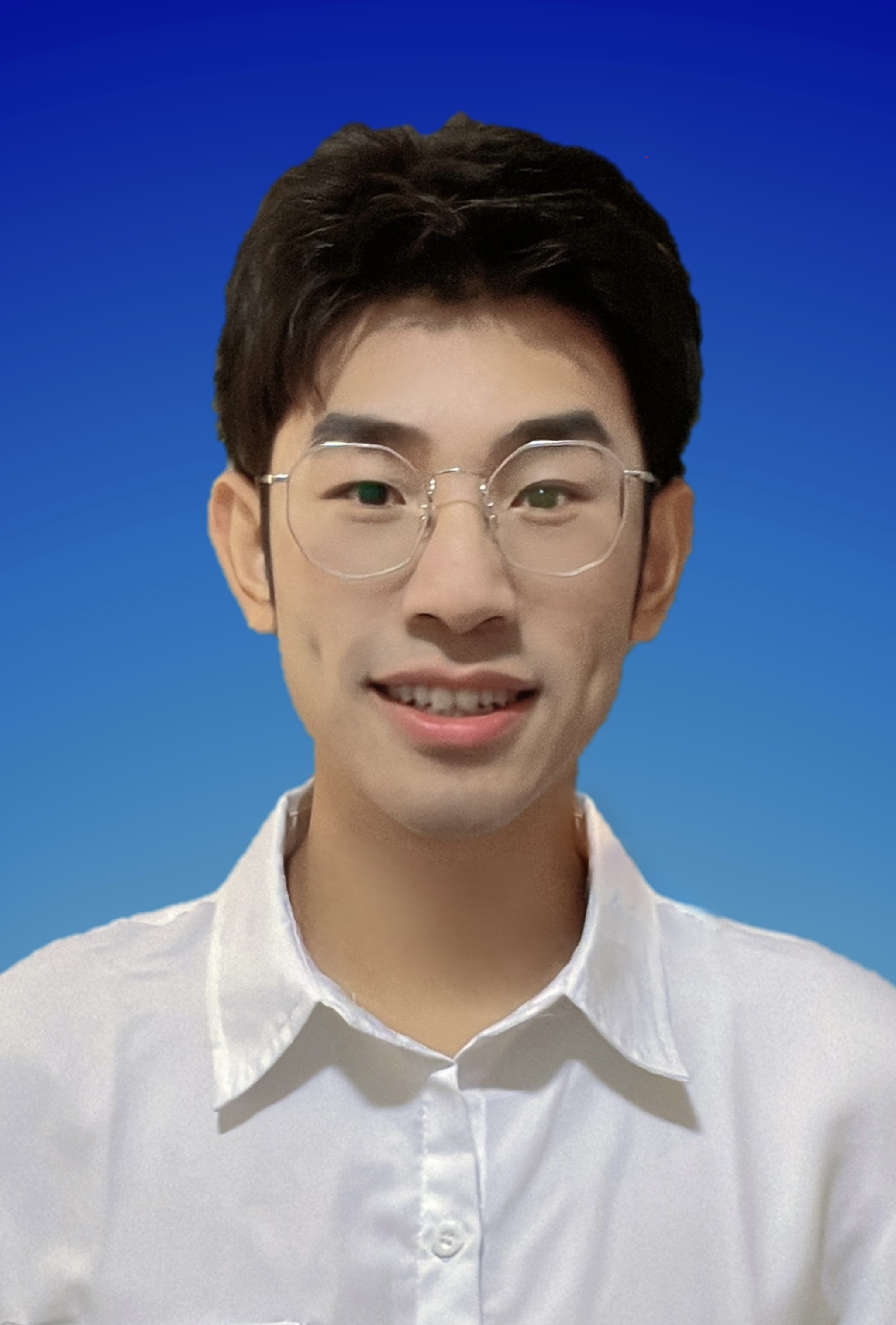}}]{Shuai Yuan}
 received the B.S. degree from Xi'an Technological University, Xi'an, China, in 2019. He is currently pursuing a Ph.D. degree at Xidian University, Xi’an, China. He is currently studying at the University of Melbourne as a visiting student, working closely with Dr. Naveed Akhtar. His research interests include infrared image understanding, remote sensing, and computer vision.
 \end{IEEEbiography}

\begin{IEEEbiography}
[{\includegraphics[width=1in,height=1.25in,clip,keepaspectratio]{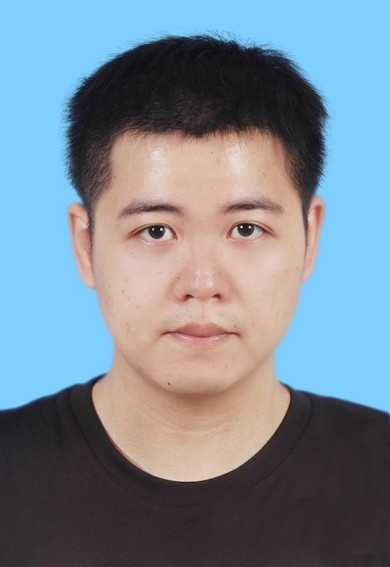}}]{Yukai Shi}
received the Ph.D. degrees from the school of Data and Computer Science, Sun Yat-sen University, Guangzhou China, in 2019. He is currently an associate professor at the School of Information Engineering, Guangdong University of Technology, China. His research interests include computer vision and machine learning.
\end{IEEEbiography}

\begin{IEEEbiography}
[{\includegraphics[width=1in,height=1.25in,clip,keepaspectratio]{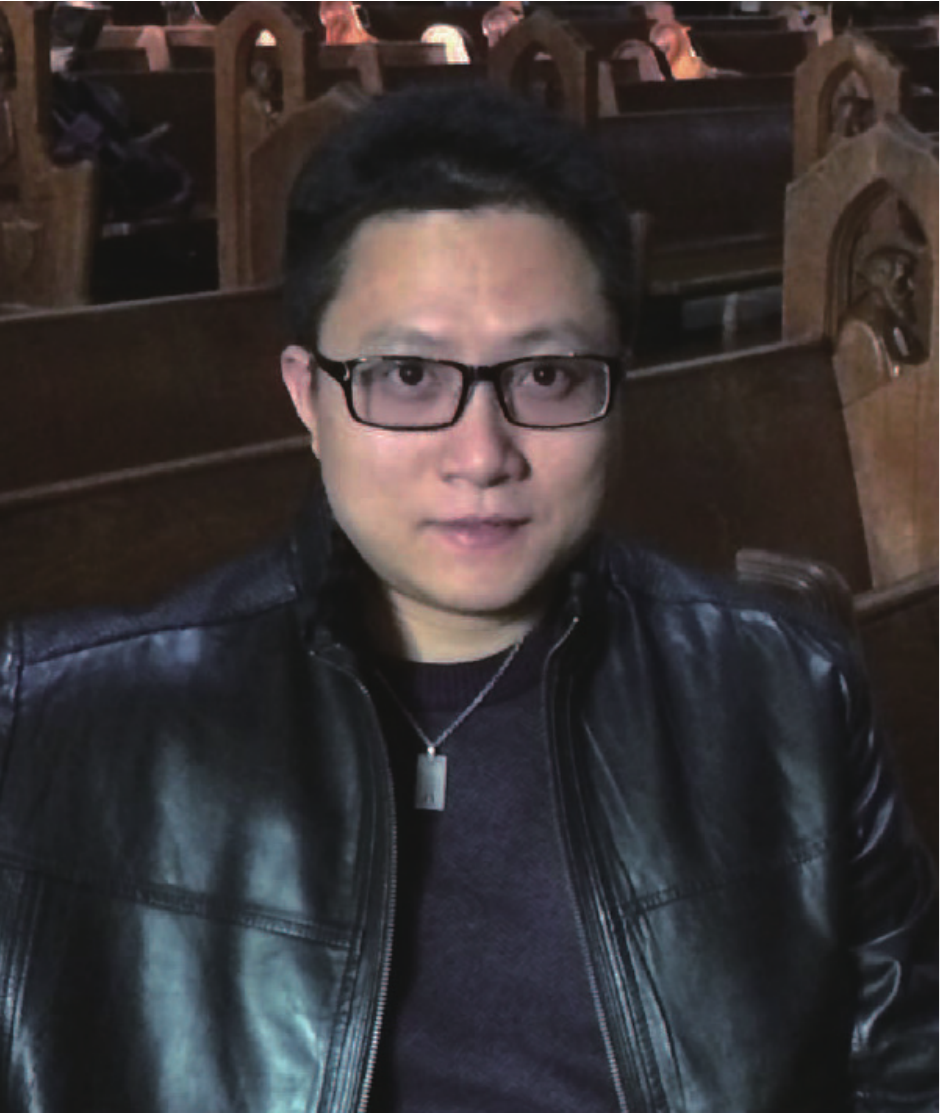}}]{Liang Lin}(Fellow, IEEE) is a Full Professor of computer science at Sun Yat-sen University. He served as the Executive Director and Distinguished Scientist of SenseTime Group from 2016 to 2018, leading the R$\&$D teams for cutting-edge technology transferring. He has authored or co-authored more than 200 papers in leading academic journals and conferences, and his papers have been cited by more than 26,000 times. He is an associate editor of IEEE Trans.Neural Networks and Learning Systems and IEEE Trans. Multimedia, and served as Area Chairs for numerous conferences such as CVPR, ICCV, SIGKDD and AAAI. He is the recipient of numerous awards and honors including Wu Wen-Jun Artificial Intelligence Award, the First Prize of China Society of Image and Graphics, ICCV Best Paper Nomination in 2019, Annual Best Paper Award by Pattern Recognition (Elsevier) in 2018, Best Paper Dimond Award in IEEE ICME 2017, Google Faculty Award in 2012. His supervised PhD students received ACM China Doctoral Dissertation Award, CCF Best Doctoral Dissertation and CAAI Best Doctoral Dissertation. He is a Fellow of IEEE/IAPR/IET.
\end{IEEEbiography}

\end{document}